%% file: Robust_PCA_3_31_2016.tex
\documentclass[journal]{IEEEtran}
\input{preamble}
\usepackage{makecell}
\begin{document}
\title{\huge{Randomized Robust Subspace Recovery for High Dimensional Data Matrices}}

\author{Mostafa~Rahmani, \IEEEmembership{Student Member,~IEEE} and George~K.~Atia,~\IEEEmembership{Member,~IEEE} % <-this % stops a space
\thanks{This material is based upon work supported by the National Science Foundation under NSF CAREER Award CCF-1552497 and NSF grant CCF-1320547.

The authors are with the Department of Electrical and Computer Engineering, University of Central Florida, Orlando,
FL 32816 USA (e-mail: mostafa@knights.ucf.edu, george.atia@ucf.edu).}% <-this % stops a spa
}

\markboth{}%
%\markboth{Journal of \LaTeX\ Class Files,~Vol.~11, No.~4, %December~2012}%
{Shell \MakeLowercase{\textit{et al.}}: Bare Demo of IEEEtran.cls for Journals}
\maketitle

\begin{abstract}
This paper explores and analyzes two randomized designs for robust Principal Component Analysis (PCA) employing low-dimensional data sketching. In one design, a data sketch is constructed using random column sampling followed by low-dimensional embedding, while in the other, sketching is based on random column and row sampling.
%, termed RED,
%constructs a data sketch using random column sampling followed by low-dimensional embedding, while the second design is based on random column and row sampling.
Both designs are shown to bring about substantial savings in complexity and memory requirements for robust subspace learning over conventional approaches that use the full scale data.
%Unlike conventional approaches that use the full scale data, robust subspace learning is applied to a significantly reduced data sketch, hence substantial savings in complexity and memory requirements are achieved.
A characterization of the sample and computational complexity of both designs is derived in the context of two distinct outlier models, namely, sparse and independent outlier models. The proposed randomized approach can provably recover the correct subspace with computational and sample complexity that are almost independent of the size of the data. The results of the mathematical analysis are confirmed through numerical simulations using both synthetic and real data.
%
%In this paper, two randomized designs for robust Principal Component Analysis (PCA) utilizing low-dimensional data sketching are investigated for distinct outlier models.
%%The randomized approach obtains a low dimensional sketch of the given data in two steps.
%In the first randomized design, a random subset of the columns of data is selected, then the selected columns are embedded into a random low dimensional subspace. In the second design, a random subset of the columns of data are sampled, then a random subset of the rows of sampled columns are selected. Contrary to the conventional approach, in the randomized approach the robust subspace learning algorithm is applied to the data sketch as apposed to the full scale data. Thus, the complexity and the memory requirement of the randomized approach are substantially less than the conventional methods. We analyze both randomized designs with different models for the outlying columns and it is shown that in each case, the randomized approach can provably recover the correct subspace with a computation and sample complexity almost independent from the size of data.
\end{abstract}

\begin{IEEEkeywords}
Low Rank Matrix, Robust PCA, Randomized Algorithm, Subspace Learning, Big Data, Outlier Detection, Sketching, Column/Row Sampling, Random Embedding
\end{IEEEkeywords}

\IEEEpeerreviewmaketitle

\section{Introduction}

%Principal Component Analysis (PCA) is a key mathematical tool used in a broad range of applications. PCA aims to approximate the data with a low-dimensional subspace.

Principal Component Analysis (PCA) has been routinely used to reduce dimensionality by finding linear projections of high-dimensional data into lower dimensional subspaces. Such linear models are highly pertinent to a broad range of data analysis problems, including computer vision, image processing, machine learning and bioinformatics \cite{lamport44,lamport45,rahmani2015innovation}.

Given a data matrix $\bD \in \mathbb{R}^{N_1 \times N_2}$, PCA finds an $r$-dimensional subspace by solving
%. Therefore, the conventional PCA problem can be written as
\begin{eqnarray}
\underset{\hat{\bU}}{\min} \| \bD - \hat{\bU} \hat{\bU}^T \bD \|_F \quad \text{subject to} \quad \hat{\bU}^T \hat{\bU} = \bI,
\label{eq1}
\end{eqnarray}
where $\hat{\bU} \in \mathbb{R}^{N_1 \times r}$ is an orthonormal basis for the $r$-dimensional subspace, $\bI$ denotes the identity matrix and $\| .
\|_F$ the Frobenius norm. While PCA is useful when the data has low intrinsic dimension, it is notoriously sensitive to outliers in the sense that the solution to % the least-square method
(\ref{eq1}) can arbitrarily deviate from the true underlying subspace if a small portion of the data is not contained in this low-dimensional subspace.

%The least-squares method is notoriously sensitive to outliers in the sense that the output of PCA can arbitrarily deviate be dramatically impacted even if just a small portion of the data does not follow the low-dimensional model.

As outliers prevail much of the real data, a large body of research has focused on developing robust PCA algorithms that are not unduly affected by the presence of outliers. The corrupted data can be expressed as
\begin{eqnarray}
\bD = \bL + \bC \: ,
\label{eq2}
\end{eqnarray}
where $\bL$ is a low rank matrix whose columns lie in a low-dimensional subspace, and the matrix $\bC$, called the outlier matrix, models the data corruption.
%the corruption from the low rank matrix. We call $\bC$ the outlier matrix.
Two main models for data corruption that are in fact incomparable for the most part were considered in the literature, namely, element-wise and column-wise corruption. In the former model, $\bC$ is an element-wise sparse matrix with arbitrary support, whose entries can have arbitrarily large magnitudes \cite{lamport1,lamport22}. In this model, all the columns of $\bL$ may be affected by the non-zero elements of $\bC$ given its arbitrary support pattern.
%There are two main models for the corruption, element-wise corruption and column-wise corruption.
%In the element-wise model, the matrix $\bC$ is an element-wise sparse matrix. This means that a small portion of the elements of $\bC$ are non-zero, but we cannot assume any structure on the pattern of the non-zero elements \cite{lamport1}. In this model, all the columns of $\bL$ can be affected by the non-zero elements of the outlier matrix $\bC$.
%
In the column-wise model, a portion of the columns of $\bC$ are non-zero and these non-zero columns do not lie in the column space of $\bL$ \cite{lamport10 , lamport24}. Thus, a portion of the columns of $\bL$, the so-called inliers, are unaffected by $\bC$. This paper focuses on the column-wise outlier model according to the following data model.
\begin{data model} The given data matrix $\bD$ satisfies the following conditions.\\
1. The matrix $\bD$ can be expressed as (\ref{eq2}).\\
2. $\rank(\bL) = r$. \\
3. The matrix $\bC$ has $K$ non-zero columns. The non-zero columns of $~\bC$
do not lie in the column space of $~\bL$. Hence, if $\mathcal{I}$ is the index set
of the non-zero columns of $\bC$ and $\bU \in \mathbb{R}^{N_1 \times r}$ an
orthonormal basis for the column space of $\bL$, then,
\begin{eqnarray}
(\bI - \bU \bU^T) \bC_i \neq 0\quad \text{for} \quad i \in \mathcal{I},
\label{eq7}
\end{eqnarray}
where $ \bC_i $ is the $i^{\text{th}}$ column of $\bC$. \\
4. Without loss of generality, it is assumed  that $$\bL_i = \mathbf{0} \quad
\text{for} \quad i \in \mathcal{I} \:,$$ where $\bL_i$  is the
$i^{\text{th}}$ column of $\bL$. Define $\bL^{'} \in \mathbb{R}^{N_1 \times
N_2^{'}}$ as the matrix of non-zero columns of $\bL$ (the inlier columns) and
$N_2^{'}$ as the number of inlier columns, i.e., $N_2 = K + N_2^{'}$.
\end{data model}

The problem of robust PCA has received considerable attention in recent years \cite{rahmani2015analysis,lamport2,lamport6,lamport10,lamport13,lamport21,lamport29,lamport47,lamport48,zhang2014novel,soltanolkotabi2012geometric}.
%While many algorithms were developed for robust PCA problem \cite{lamport2,lamport6,lamport10,lamport13,lamport21,lamport29,lamport47,lamport48,zhang2014novel,soltanolkotabi2012geometric}.
However, the state-of-the-art robust estimators and matrix decomposition techniques are mostly unscalable, which limits their usefulness in big data applications.
%However, the vast majority of existing algorithms are not directly applicable in high-dimensional settings due to their computational complexity and memory requirements.
%In addition, many of them are not applicable to high dimensional data applications because they impose intolerable computation complexity and memory requirement.
For instance, many of the existing approaches rely on iterative algorithms that involve computing a Singular Value Decomposition (SVD) of the $N_1\times N_2$ data matrix in each iteration, which is computationally prohibitive in high-dimensional settings. This motivates the work of this paper.
%For example, many of the existing methods perform Singular Value Decomposition (SVD) of $N_1 \times N_2$ matrices at each iteration, which is a computationally expensive operation for high-dimensional data matrices.

\subsection{Notation and definitions}
 Given a matrix $\bL$, $\|\bL\|$ denotes its spectral norm, $\| \bL \|_{*}$ its nuclear norm which is the sum of the singular values, and $\| \bL \|_1$ its $\ell_1$-norm given by $ \| \bL \|_1 = \sum\limits_{i,j} \big | \bL (i,j) \big |$, i.e. the sum of the absolute values of its entries.
%\label{eq3}
%\end{eqnarray}
The norm $\| \bL \|_{1,2}$ is defined as
%\begin{eqnarray}
$\| \bL \|_{1,2} = \sum\limits_{i}  \| \bL_i \|_2, $
%\label{eq4}
%\end{eqnarray}
where $\| \bL_i \|_2$ is the $\ell_2$-norm of the $i^{\text{th}}$ column of
$\bL$. In an $N$-dimensional space, $\be_i$ is the $i^{\text{th}}$ vector of
the standard basis. For a given vector $\ba$, $\| \ba \|_p$ denotes its $\ell_p$-norm.
Two linear subspaces $\calS_1$ and $\calS_2$ are said to be independent if the dimension of their intersection $\calS_1 \cap \calS_2$ is equal to zero.
%For a given matrix $\bA$,- $\bV_A$ is the matrix of right singular vectors of $\bA$ whose span is equal to the row space of $\bA$.
In the presented algorithms and analysis, we make use of the following definitions.
\begin{definition}
The row space of a matrix $\bL$ with rank $r$ and $N_2^{'}$ non-zero columns is said to be incoherent with parameters $\mu_v$, $\eta_v$ and $\gamma$ if
\begin{eqnarray}
\begin{aligned}
&\quad \underset{i}{\max} \|\bV^T \be_i\|_2^2 \leq \frac{r \mu_v}{N_2} \quad , \quad \eta_v \hspace{-0.5mm}= \hspace{-.5mm}\sqrt{N_2}
\max_{i,j} |\bV(i,j)|  \: , \\
& \quad \quad \quad \quad \quad \quad \quad  \underset{i}{\max} \|\bV^T \be_i\|_2^2 \leq \frac{r \gamma}{N_2^{'}}
\end{aligned}
\end{eqnarray}
where $\bV$ is an orthonormal basis for the row space of  $\bL$. Similarly, the column space of $\bL$  is said to be incoherent with parameters $\mu_u$ and $\eta_u$ if
\begin{eqnarray}
 \underset{i}{\max} \|\bU^T \be_i\|_2^2 \leq \frac{r \mu_u}{N_1}  \quad and \quad \eta_u \hspace{-0.5mm}= \hspace{-.5mm}\sqrt{N_1}
\max_{i,j} |\bU(i,j)| \:.
\end{eqnarray}
\label{def: incoh}
\end{definition}

\begin{definition}
(Distributional Johnson-Lindenstrauss (JL) Property \cite{ lamport35 , lamport37 , lamport39}). An $m \times n$ matrix $\mathbf{\Phi}$ is said to satisfy the Distributional JL property if for any fixed $\bv \in \mathbb{R}^{n}$ and any $\epsilon \in (0,1)$,
\label{def2}
\begin{eqnarray}
\mathbb{P} \left( \Big | \| \mathbf{\Phi} \bv \|_2^2 - \| \bv \|_2^2 \Big | \ge \epsilon  \| \bv \|_2^2  \right) \leq 2 e^{- m f(\epsilon) },
\label{eq12}
\end{eqnarray}
where $f(\epsilon) > 0$ is a constant that is specific to the distribution of $\mathbf{\Phi}$ and depends only on $\epsilon$.
\label{def: embedd}
\end{definition}
We refer the reader to \cite{lamport49,candes2007sparsity} for further details concerning the properties of the incoherency parameters.
Also, similar to Definition \ref{def: incoh}, we define $\mu_v^{'}$ as the row space incoherency of $\bL^{'}$, i.e., if $\bV^{'} \in \mathbb{R}^{N_2^{'} \times r}$ is an orthonormal basis for the row space of $\bL^{'}$, $\underset{i}{\max} \| \be_i^T \bV^{'} \|_2^2 \leq \frac{r \mu_v^{'}}{N_2^{'}}$.

\subsection{Summary of contributions}
Motivated by the aforementioned limitation of existing approaches in big data settings, which is to be further elaborated in the related work section, this paper explores and analyzes a randomized approach to robust PCA using low-dimensional data sketching. Two randomized designs are considered.
%his paper, we investigate two randomized designs for the robust PCA problem.
The first design is the Random Embedding Design (RED) wherein a random subset of the data columns is selected then embedded into a random low-dimensional subspace. The second randomized design is a Random Row-sampling Design (RRD), in which a random subset of the data columns are sampled, then we select a random subset of the rows of the sampled columns. Unlike conventional robust PCA algorithms that use the full-scale data, robust subspace recovery is applied to the reduced data sketch.
%In contrary to the conventional robust PCA algorithms, the robust subspace recovery algorithm is applied to the data sketch as apposed to the full-scale data.

We consider two distinct popular models for the outlier matrix. In the first model -- the independent outlier model -- it is assumed that any small subset of the non-zero columns of $\bC$ is not linearly dependent. This model allows for a remarkable portion of the data to be outliers. In the second model -- the sparse outlier model -- it is assumed that $\bC$ is column-sparse, i.e., a very small portion of the given data columns are outliers, but no assumption is made about the linear dependence of the outlying columns.
%this model does not make any assumption about linear dependency of the outlying columns.
For  both outlier models, we prove that the randomized approach using either of the designs can recover the correct subspace with high probability (whp). Some of the key technical contributions of this paper are listed below. \\
1. To the best of our knowledge, RRD is used and analyzed here for the first time for robust PCA with column-wise corruption. We prove that RRD can recover the correct subspace using roughly $\calO (r^2 \mu_u \mu_v)$ random linear data observations. %The data sketching step of the RRD impose no computational complexity and
The complexity of subspace recovery in RRD is roughly $\calO (r^3 \mu_u \mu_v)$.
 \\
2. For RED, it is shown here for the first time that the sufficient number of random linear data observations for correct subspace recovery is roughly $\calO (r^2 \mu_v)$. \\
3. The proposed randomized approach based on the linear independence of the outlier columns is novel. We take advantage of random column sampling to substantially reduce the number of outlying columns. Thus, unlike conventional approaches that need to go through all the columns to identify the outliers, we only need to check $\calO (r \mu_v)$ data points. \\
Table \ref{tabl 1} summarizes the derived order of sufficient number of linear random data observations for the randomized designs with both outlier models.

\begin{table}
\centering
\caption{Order of sufficient number of random linear data observations.}
\begin{tabular}{| c | c | c |}
\hline
Outlier Model/Design & RED & RRD \\
 \hline
 Column-sparsity & $r^2  \max (\mu_v , r \mu_v^2 \: {K}/{N_2} )$ & $r^2 \eta_u^2 \max (\mu_v , r \mu_v^2 \: {K}/{N_2} )$ \\
 \hline
 Independence & $r^2 \mu_v^{'} \max (1 , \mu_v^{'} \: K/N_2) $ & $r^2 \mu_v^{'} \max (\eta_u^2 , \mu_v^{'} \: K/N_2) $ \\
 \hline
\end{tabular}
\label{tabl 1}
\end{table}

%In both structures, it is proved that the randomized strcutures can correctly recover the subspace and locate the outlying columns with a complexity almost independent from the size of data. The innovations of this paper can be categorized as follows.

 % In the presented randomized structure, first a small subset of the columns of $\bD$ are selected randomly and then the selected columns are embedded into a random low dimensional subspace. The robust subspace learning algorithm is applied to this small size compressed data sketch.
%We consider different popular models for the outlying data and it is shown that with appropriate robust subspace learning algorithm, the randomized structure can correctly recover the subspace with a complexity almost independent from the size of the full-scale data. To the best of our knowledge, it is shown here for the first time that the randomized structure  provably recovers the correct subspace (i.e., identify outlying columns exactly) with high probability using roughly $\calO (r^2 \mu_v)$ where $\mu_v$ is the row-space coherency parameter of $\bL$ (Definition \ref{def5}). In addition, we study the limitation of uniformly random column sampling from structured low rank matrices and proposed some techniques to achieve more efficient sampling procedure.

\section{Related Work}
\subsection{Robust PCA}
%Due to the importance of PCA and outlier detection in many applications, there has been many works focusing on the robust PCA problem.
Some of the earliest approaches to robust PCA relied on robust estimation of the data covariance matrix, such as S-estimators, the minimum covariance determinant,
the minimum volume ellipsoid, and the Stahel-Donoho estimator \cite{lamport47}. However, these approaches are not applicable in high-dimensional settings due to their computational complexity and memory requirements, and we are not aware of any scalable algorithms for implementing these methods with explicit performance guarantees.

Another popular approach replaces the Frobenius norm in (\ref{eq1}) with other norms to enhance robustness to outliers \cite{lamport18}. An instance of this approach is \cite{lamport29}, which uses an $\ell_1$-norm relaxation, commonly used for sparse vector estimation, yielding robustness to outliers \cite{decod,lamport25,lamport22}. \cite{lamport21} replaces the $\ell_1$-norm in \cite{lamport29} with the $\ell_{1,2}$-norm to promote column sparse solutions. %The work in \cite{lamport21} modifies \cite{lamport29} by exploiting the column sparse structure of the outlier matrix (the column sparsity of $\bC$).
%However, these approaches are not directly applicable to large data matrices. %However, these methods are computationally expensive for large data matrices and there are no guarantees on their performance.
Recently, the idea of using a robust norm was revisited in \cite{lamport2, zhang2014novel}. Therein, the non-convex constraint set is relaxed to a larger convex set and exact subspace recovery is guaranteed under certain conditions.
%the hard optimization problem (\ref{eq1}) is relaxed by replacing the non-convex constraint set with a larger convex set. Under certain conditions, it is shown that the exact subspace can be extracted.
Nevertheless, these approaches are not directly applicable to large matrices and high-dimensional data settings. For example, the iterative solver of \cite{lamport2} requires an eigen-decomposition operation in each iteration. %In addition, the assumption of \cite{zhang2014novel} is rather restrictive because it assumes that the columns of $\bD$ span the $N_1$-dimensional ambient space.

An interesting approach for outlier detection was recently proposed in \cite{lamport8,soltanolkotabi2012geometric} based on the idea that outliers do not typically follow low dimensional structures. Hence, few outliers cannot form a linearly dependent set. %unlike the inliers which lie in a low dimensional subspace, a small number of them  are linearly dependent.
%This feature is leveraged effectively to distinguish the outlier data points in  \cite{lamport8,soltanolkotabi2012geometric}.
Unlike most existing approaches, this approach can recover the correct subspace even if a remarkable portion of the data is outliers, albeit its computational complexity is roughly $\calO (N_2^2)$ \cite{soltanolkotabi2012geometric}. Also, the number of samples in \cite{lamport8} scales linearly with the data dimension, which is quite restraining in high dimensional settings.
%However, this approach need to save the whole data in the working memory and the computation complexity for subspace recovery is roughly $\calO (N_1 N_2)$.

In this paper, we propose two randomized algorithms for two distinct outlier models. The first algorithm is a new randomized approach that exploits the linear independence of the outlying columns (cf. Section \ref{sec:algo1_ind_outliers}). It is shown that this randomized algorithm can recover the correct subspace with sample complexity that is almost independent of the size of the data (roughly $\calO (r^2 \mu_v)$). It also imposes less stringent constraints on the distribution of outliers compared to prior work on the independent outlier model.
%In the methods based on this idea, all the columns should be checked to find the outliers. In this paper, we show that using the randomized structure, a small subset of the columns are enough to check and the randomized approach requires a weaker assumption on the distribution of the outlying columns.

%Recently, much research was devoted to developing convex
%algorithms for low rank matrix recovery.
The second algorithm presented in Section \ref{sec:algo2_sparse_outliers} deals with the sparse column outlier model using convex rank minimization on reduced data sketches. Robust PCA using convex rank minimization \cite{lamport34} was first analyzed in
\cite{lamport1,lamport10}. It was shown in \cite{lamport10} that the optimal point of \begin{eqnarray}
\begin{aligned}
& \underset{\hat{\bL},\hat{\bC}}{\min}
& & \|\hat{\bL} \|_* + \lambda\|\hat{\bC}\|_{1,2} \\
& \text{subject to}
& & \hat{\bL} + \hat{\bC} = \bD \:.
\end{aligned}
\label{eq6}
\end{eqnarray}
yields the exact subspace and the correct outliers identification provided that $\bC$ is sufficiently column-sparse. The column-sparsity of $\bC$ is the main requirement of (\ref{eq6}), i.e., a very small fraction of the columns of $\bC$ can be non-zero. The computational complexity of (\ref{eq6}) is roughly $\calO (r N_1 N_2)$ per iteration and the entire data needs to be saved in the working memory, which is prohibitive in big data applications. In this paper, we show that the complexity of subspace recovery reduces to $\calO (r^3 \mu_v)$, which is  substantially less than $\calO (r N_1 N_2)$ for high dimensional data, using a randomized approach that applies (\ref{eq6}) to reduced data sketches.

%Algorithm (\ref{eq6}) is not applicable to big data since it impose

%It was shown that (\ref{eq6}) yields a correct decomposition if the fraction of non-zero columns of $\bC$ is less than a constant threshold \cite{lamport10}.

\subsection{Randomized approaches for Robust PCA}
The low rank component $\bL$ has a low-dimensional structure, and so is $\bC$ in the element-wise sparse or column-wise sparse models. These low-dimensional structures motivated the usage of randomized algorithms for robust PCA using small sketches constructed from random linear measurements of $\bD$.
%a series of research works to solve the robust PCA problem using a small set of random linear measurements of $\bD$.

However, the  majority of such algorithms have focused on robust PCA with the element-wise outlier model \cite{lamport28,myp,liu2011solving,mackey2011divide,mu2011accelerated,zhou2011godec}.  For instance, two randomized methods were proposed in \cite{mackey2011divide} and \cite{myp} to recover $\bL$ from small subsets of the columns and rows of $\bD$. The randomized approach in \cite{myp} was shown to reduce complexity from $\calO (N_1 N_2 r)$ to $\calO(\max (N_1 , N_2) r^2)$ per iteration.

%The low rank and the sparse matrix follow low-dimension geometrical structures, which provides the opportunity to solve the robust PCA problem using a small set of random observations of the data \cite{lamport28,myp,vay}.
%The low rank matrix can be written as $\bL = \bU \bQ^T$ where $\bU \in \mathbb{R}^{N_1 \times r}$ and $\bU \in \mathbb{R}^{N_2 \times r}$. The matrix $\bU$ is a basis for the columns subspace and $\bQ$ is a basis for the rows subspace of $\bL$. An interesting approach is to use randomized techniques to learn the columns and rows subspace. These subspaces are low dimensional, thus they can be obtained from a small subset of given data \cite{myp}. Given that, not only is the robust PCA problem translated to low-dimensional optimization problems (subspace learning problem), but also a small set of random linear measurements of the data is used instead of the whole data. In \cite{myp , vay}, it was shown that under the first outlier model, the columns subspace of the low rank matrix can be learned from a small random subset of the columns of the data matrix and the rows subspace can be obtained from a small random subset of the rows of the data matrix. Roughly speaking, \cite{myp} and \cite{vay} solve the robust PCA problem (under the first outlier model) using $\mathcal{O}(r \log N_2)$ randomly sampled columns and $\mathcal{O} (r \log N_1)$ randomly sampled rows. They solve the robust PCA problem (with the first outlier model) with the same number of linear observations as in \cite{lamport28}, but with significantly faster algorithms.
Randomized approaches for the column-wise outlier model were proposed in \cite{lamport9} and \cite{lamport8}. The algorithm in \cite{lamport8} is built on the assumption that any subset of outlying columns with cardinality less than $N_1$ is linearly independent. The algorithm repeatedly samples $N_1$ data points until a linearly dependent set is found, upon which those columns that do not depend linearly on the other ones are selected as outliers. Since the number of samples scales linearly with the data dimension and the algorithm requires $\calO(N_1^2 N_2)$ iterations on average, it can be quite restraining in high dimensions, especially when a remarkable portion of the data is ouliers.
%In this algorithm, $N_1$ columns are randomly sampled, and those columns that are not linearly dependent on the other columns are selected as outliers. The number of sampled columns scales with the data dimension which may be quite restraining for high dimensional data.
%In addition, the analysis presented in \cite{lamport8} requires too many iteration when a remarkable portion of data is oulier.
Another limitation of \cite{lamport8} emerges from the assumption that any subset of inliers with at least $r$ columns spans the column space of $\bL$. This may not be true in general, especially with real world data which often exhibits clustering structures.
%Another limitation of the method proposed in \cite{lamport8} is that it fails when some of the selected outliers are not linearly independent. To dispense with these limitations, in this paper we show that the required number of randomly selected columns is independent of the dimension of the given data, and propose an algorithm that is robust to linearly dependent outliers. In the proposed idea, the robust PCA problem is interpreted as a subspace clustering problem.

%In \cite{lamport9}, a randomized algorithm for the robust PCA problem was presented.
The work in \cite{lamport9} considers the column-sparse outlier model. The data is first embedded into a random low-dimensional subspace, then a subset of the columns of the compressed data is selected. The convex program in  (\ref{eq6}) is then used to locate the outlying columns of the compressed data. The analysis provided in \cite{lamport9} requires roughly $\calO(r N_2)$ random linear observations for exact outlier detection. In this paper, we show that both the required number of sampled columns and the dimension of the subspace for random embedding are almost independent of the size of data and the required number of random linear measurements is shown to be roughly $\calO(r^2 \mu_v)$. %In addition, we consider other model assumptions for $\bC$ (other than column-sparsity) and it is shown that for each model the randomized approach can guarantee the exact subspace recovery with high probability.

\section{Proposed Approach}
\label{sec:prop_approach}
In this section, we propose two algorithms for two distinct models of the outlier matrix. In the first model, the independent outlier model, it is assumed that any small subset of outliers is not linearly dependent. The corresponding algorithm is easy to implement and can recover the correct subspace even if more than 90$\%$ of the data is outliers. The second model concerns the scenario in which $\bC$ is column-sparse, yet allows for outliers to be linearly dependent. For both algorithms, we consider two randomized designs, one utilizing random embedding and the other using random row sampling. We provide a full analysis of the sample complexity for the two algorithms based on both randomized designs. The randomized algorithms can provably retrieve the correct subspace with computational and sample complexity that are almost independent of the size of $\bD$. In this section, we present the algorithms and the key insights underlying the proposed approach along with the statement of the main theorems. A step-by-step analysis is deferred to Sections \ref{sec:analysis 1} and \ref{sec:analysis 2}.
% In addition, we discuss a more efficient column sampling method.

\begin{algorithm}
\caption{Randomized Robust PCA based on outlier linear independence with both randomized designs}
{\footnotesize
\textbf{Input}: Data matrix $\bD \in \mathbb{R}^{N_1 \times N_2} $
%\textbf{1. Initialization.} Form column sampling matrix $\bS \in \mathbb{R}^{N_2 \times m_1}$ and random embedding matrix $\mathbf{\Phi} \in \mathbb{R}^{m_2 \times N_1} $.

\smallbreak
\textbf{1. Data Sketching}\\
\textbf{1.1} {\it Column Sampling:}
 Matrix $\bS \in \mathbb{R}^{N_2 \times m_1}$ samples $m_1$ columns of $\bD$ randomly $\bD_{s} = \bD \bS$. The columns of $\bS$ are a set of standard basis vectors. Thus, $\bD_s \in \mathbb{R}^{N_1 \times m_1}$.

\textbf{1.2} {\it Row Compression:}\\
\textcolor{blue}{\textbf{If we use RED:}} Matrix $\mathbf{\Phi} \in \mathbb{R}^{m_2 \times N_1}$ is drawn from any distribution satisfying (\ref{eq12}).
Matrix $\mathbf{\Phi}$ projects the sampled columns $\bD_s$  into a random $m_2$-dimensional subspace $\bD_s^{\phi}=\mathbf{\Phi} \bD_s$. Thus, $\bD_s^{\phi} \in \mathbb{R}^{m_2 \times m_1}$.\\
\textcolor{blue}{\textbf{If we use RRD:}} The rows of $\mathbf{\Phi} \in \mathbb{R}^{m_2 \times N_1}$ are a subset of standard basis. Matrix $\mathbf{\Phi}$ samples $m_2$ rows of sampled columns $\bD_s^{\phi}=\mathbf{\Phi} \bD_s$. Thus, $\bD_s^{\phi} \in \mathbb{R}^{m_2 \times m_1}$.

\smallbreak
\textbf{2. Subspace Learning}\\
\textbf{2.1} {\it Sampled Outlier Columns Detection:} Define $\bd_{\phi s}^i$ as the $i^{\text{th}}$ column of $\bD_s^{\phi}$ and $\bQ_i^{\phi}$ is equal to $\bD_s^{\Phi}$ with the $i^{\text{th}}$ column removed.
Solve the optimization problem (\ref{eq13}) %for the columns of $\bD_s^{\phi}$
for $1 \leq i \leq m_1$ to identify the outlying columns of $\bD_s^{\phi}$ (if the minimum value of (\ref{eq13}) is non-zero, the column is an outlier). \\
\textbf{2.2} Subspace Learning: Construct $\bT$ as the set of columns of $\bD_s$ corresponding to a set of linearly independent inlier columns of $\bD_s^{\phi}$ spanning the subspace of the inlier columns of $\bD_s^{\phi}$.
\smallbreak
\textbf{Output:} The matrix $\bT$ is a basis for the column space of $\bL$.
}
\end{algorithm}

\begin{algorithm}
\caption{ Randomized Robust PCA based on outlier matrix column-sparsity with both randomized designs}
{\footnotesize
\textbf{Input}: Data matrix $\bD \in \mathbb{R}^{N_1 \times N_2} $
%\textbf{1. Initialization.} Form column sampling matrix $\bS \in \mathbb{R}^{N_2 \times m_1}$ and random embedding matrix $\mathbf{\Phi} \in \mathbb{R}^{m_2 \times N_1} $.

\smallbreak
\textbf{1. Data Sketching}\\
Perform steps 1.1 and 1.2 of Algorithm 1.

\smallbreak
\textbf{2. Subspace Learning}\\
\textbf{2.1} {\it Sampled Outlier Columns Detection:} Obtain $\hat{\bL}_s^{\phi}$ and $\hat{\bC}_s^{\phi}$ as the optimal solution of
\begin{eqnarray}
\begin{aligned}
& \underset{\dot{\bL}^{\phi}_s,\dot{\bC}^{\phi}_s}{\min}
& & \lambda\|\dot{\bC}^{\phi}_s\|_{1,2}  + \|\dot{\bL}^{\phi}_s \|_* \\
& \text{subject to}
& & \dot{\bL}^{\phi}_s +\dot{\bC}^{\phi}_s = \bD^{\phi}_s \: .
\end{aligned}
\label{convex_in_alg}
\end{eqnarray}
The non-zero columns of $\hat{\bC}_s^{\phi}$ indicate the location of the outlying columns.  \\
\textbf{2.2} Subspace Learning: Construct $\bT$ as the set of columns of $\bD_s$ corresponding to a set of linearly independent inliers of $\bD_s^{\phi}$ spanning the subspace of the inlier columns of $\bD_s^{\phi}$.
\smallbreak
\textbf{Output:} The matrix $\bT$ is a basis for the column space of $\bL$.
}
\end{algorithm}

\subsection{Algorithm 1: randomized approach for the independent outlier model}
\label{sec:algo1_ind_outliers}
Algorithm 1 hinges on the assumption that any small subset of outliers are linearly independent as stated next.
%More specifically, it is assumed that any subset of the non-zero columns of $\bC$ with cardinality less than or equal to $\bq$ are linearly independent.
%
\begin{assumption}
Any subset of the non-zero columns of $\bC$ with cardinality equal to $q$ spans a $q$-dimensional subspace that is independent of the column space of $\bL$.
\label{asm1}
\end{assumption}
The requirement on $q$ will be formalized later in the section. The table of Algorithm 1 presents the algorithm with both randomized designs along with the definitions of the used symbols. The only difference is in step 1.2 as RED uses random embedding while RRD uses row sampling.

\noindent{\it Insight}: Suppose that $n_s$ columns sampled randomly from $\bL^{'}$ span its column space whp.
%if $n_s$ columns of $\bL^{'}$ are sampled randomly, the selected columns span the column space of $\bL^{'}$ whp.
We do not have direct access to $\bL^{'}$ but assume that the number of sampled data columns, $m_1$, is large enough so that the number of inliers in $\bD_s$ (the sampled data columns) is at least $(n_s+1)$ and the number of outliers is less than $q$ whp. In Section \ref{sec:analysis 1}, it is shown that the sufficient values for $n_s$, $m_1$ and the upper-bound $q$ are small and scale linearly with $r$.

%Algorithm 1 is based on this idea that while the inlier columns lie in a low dimensional subspace, any small subset of outlying columns are not linearly dependent. More specifically, as it is detailed in the following assumption, it is assumed that any subset of the non-zero columns of $\bC$ with cardinality smaller than $\bq$ are linearly independent.
%\begin{assumption}
%%Suppose $\bL = \bU \mathbf{\Sigma} \bV^T$ is the compact SVD of $\bL$ where $\bU \in \mathbb{R}^{N_1 \times r}$, $\bV \in \mathbb{R}^{N_2 \times r}$ and $\mathbf{\Sigma} \in \mathbb{R}^{r \times r}$. Define $\bC_p = (\bI - \bU \bU^T)\bC$ which is the projection of the outlier matrix on the complement subspace of the column space of $\bL$.
%Any subset of the non-zero columns of $\bC$ with cardinality  equal to $q$ span a $q$-dimensional subspace which is independent from the column space of $\bL$.
%\label{asm1}
%\end{assumption}
According to Assumption \ref{asm1}, if ${\bd}_s^i$ (the $i^{\text{th}}$ column of $\bD_s$) is an inlier, then it must lie in the span of the other columns of $\bD_s$ which contains at least $(n_s + 1)$ inliers. By contrast, if ${\bd}_s^i$ is an outlier, it would not lie in the span of the other columns since the selected outliers are not linearly dependent.
%But, since the selected outliers are not linearly dependent, if ${\bd}_s^i$ is an outlying column, it does not lie in the span of the other columns of $\bD$.
This is the basis for locating the outlying columns of $\bD_s$.
%We  utilize this feature to locate the outlying columns of $\bD_s$.

Algorithm 1 solves a low-dimensional outlier identification problem by projecting the sampled data $\bD_s$ in a lower-dimensional subspace. Specifically, we form the compressed matrix \begin{eqnarray}
\bD_s^{\phi} = \mathbf{\Phi} \bD_s,
\label{eq11}
\end{eqnarray}
where $\mathbf{\Phi} \in \mathbb{R}^{m_2 \times N_1}$. The randomized designs differ in the choice of $\mathbf{\Phi}$ in (\ref{eq11}). Specifically, in RED the matrix $\mathbf{\Phi}$ embeds the sampled columns into a random low dimensional subspace, while in RRD $\mathbf{\Phi}$ samples a random subset of the rows of $\bD_s$ (c.f. table 1).

%The inlier columns of $\bD_s$ span a low dimensional subspace. Also, Usually, there are few outlying columns in $\bD_s$. Therefore, they also span a low dimensional subspace in the $N_1$- dimensional space. Thus, we project the sampled data into a  low dimensional subspace and solve the outlier detection problem there. There is an important difference between the randomized designs. In the RED, we embed the sampled columns into a random low dimensional subspace, but, in the RRD, a random subset of the rows of $\bD_s$ are sampled.  The compressed matrix can be written as
%\begin{eqnarray}
%\bD_s^{\phi} = \mathbf{\Phi} \bD_s,
%\label{eq11}
%\end{eqnarray}
%where $\mathbf{\Phi} \in \mathbb{R}^{m_2 \times N_1}$.
In order to ensure that (\ref{eq11}) preserves the essential information, we derive sufficient conditions to satisfy the following requirement.
%a sufficient value for $m_2$ to guarantee that the following requirements are complied.
\begin{requirement}
The data sketching has to ensure that:\\
\textbf{1.} The rank of $ \mathbf{ \Phi} \bL$ is equal to $r$.\\
\textbf{2.} The non-zero columns of $\mathbf{\Phi} \bC_s$ are independent and they span a subspace independent from the column space of $ \mathbf{ \Phi} \bL$.
\end{requirement}

 Define $\bd_{\phi s}^i$ as the $i^{\text{th}}$ column of $\bD_s^{\phi}$ and $\bQ_i^{\phi}$ is equal to $\bD_s^{\phi}$ with the $i^{\text{th}}$ column removed. In order to locate the outlying columns of $\bD_s^{\phi}$, we solve
\begin{eqnarray}
\underset{\hat{\bz}}{\min} \quad \| \bd_{\phi s}^i -  \bQ_i^{\phi} \hat{\bz} \|_2 \:,
\label{eq13}
\end{eqnarray}
for $1 \leq i \leq m_1$. If the minimum of (\ref{eq13}) is zero (or close to zero for noisy data) for the $i^{\text{th}}$ column, it is concluded that the $i^{\text{th}}$ column is an inlier, otherwise it is identified as an outlier.
Once the outlying columns of $\bD_s^{\phi}$ are detected, we can estimate the dimension of the subspace spanned by the inliers of $\bD_s^{\phi}$. If the estimated dimension is equal to $\hat{r}$, we find $\hat{r}$ independent inlier columns of $\bD_s^{\phi}$. Define $\bT$ as the matrix formed from the $\hat{r}$ columns of $\bD_s$ corresponding to these $\hat{r}$ independent inliers of $\bD_s^{\phi}$. Thus, if the outlying columns of $\bD_s^{\phi}$ are correctly located, $\bT$ would be a basis for the column space of $\bL$.

In many applications, we may also be interested in locating the outlying columns. If $\bT$ spans the column space of $\bL$, we can easily identify the non-zero columns of $\bC$ as the non-zero columns of
$
(\bI - \bT (\bT^T \bT)^{-1} \bT^T)\bD \:.
$
If outlier detection is intended, an alternative course for data sketching would be to start with row compression followed by column sampling. This is particularly useful in a distributed network setting, in which each agent sends a compressed version of its data vector to a central processor as opposed to centralizing the entire data. As such, the central unit would work with $\bD^{\phi} = \mathbf{\Phi} \bD$. A random subset of the columns of $\bD^{\phi}$ is then sampled to form $\bD_s^{\phi}$, and subspace learning is applied to $\bD_s^{\phi}$ to learn the column space of $\mathbf{\Phi} \bL$.
%An alternative sequence for the data sketching step of the randomized approach is to do first the row compression, then the column sampling. For instance, if we are working in a network setting, all the information does not need to be sent to the central processor and each agent sends a compressed version of its data vector. Thus, the processor is working with $\bD^{\phi} = \mathbf{\Phi} \bD$. A random subset of the columns of $\bD^{\phi}$ is sampled to form $\bD_s^{\phi}$. The subspace learning step of the randomized approach is applied to  $\bD_s^{\phi}$ to learn the column space of $\mathbf{\Phi} \bL$.
If $\hat{\bU}^{\phi}$ denotes the obtained orthonormal basis for the column space of $\mathbf{\Phi} \bL$, then the non-zero columns of $\bC$ are identified as the non-zero columns of
\begin{eqnarray}
 \bH =  (\bI - \hat{\bU}^{\phi} (\hat{\bU}^{\phi})^T)\bD^{\phi} \:.
\label{outlier_detect 1}
\end{eqnarray}
We can readily state the following theorems, which establish performance guarantees for Algorithm 1 with both randomized designs.

\begin{theorem}[Sufficient Condition-Algorithm 1 with RED]
%\label{thm:alg_1}
Suppose $\bD$ follows Data Model 1, Assumption 1 is satisfied,  $m_1$ columns are sampled randomly with replacement and any repeated columns are removed. If for fixed $c > 1$ and small $0 < \delta \ll 1/5$,
\begin{eqnarray}
\begin{aligned}
& m_1 \ge \beta \alpha \frac{N_2}{N_2^{'}} \:, \quad\beta \ge 2 + \frac{3}{\alpha} \log \frac{2}{\delta}\: , \quad q = \alpha \left( \frac{\beta K}{N_2^{'}} + \frac{1}{c} \right),\\
&m_2 \ge \max \bigg[ \frac{(r+q) \log (42 \sqrt{2}) + \log \frac{2}{\delta}}{f (\frac{1}{{2}})} \: ,\\
&\quad \quad \quad\quad \quad\frac{(r+1) \log (42 \sqrt{2}) + \log K + \log \frac{2}{\delta}}{f (\frac{1}{{2}})}  \bigg]
\end{aligned}
\label{suff thorem 1}
\end{eqnarray}
where the embedding $m_2\times N_1$ matrix $\mathbf{\Phi}$ is drawn from any
 distribution satisfying (\ref{eq12}) and
\begin{eqnarray}
\alpha = \max \Big \{ 20 \mu_v^{'} r  \log \frac{4 r}{\delta} , 3 c^2 \frac{K}{N_2^{'}} \log \frac{2}{\delta} \Big \},
\label{alpha eq}
\end{eqnarray}
then Algorithm 1 with RED yields the exact subspace and (\ref{outlier_detect 1}) identifies the non-zero columns of $\bC$ correctly
with probability at least $1 - 5 \delta$.
\label{theorem 1}
\end{theorem}
The following theorem is the counterpart of Theorem \ref{theorem 1} with RRD. %establishes sufficient conditions for Algorithm 1 with RRD.
In this paper, for the analysis of RRD we assume that the non-zero entries of $\bC$ are sampled from a zero-mean normal distribution.

\begin{theorem}[Sufficient Condition- Algorithm 1 with RRD]
%\label{thm:alg_1}
Suppose $\bD$ follows Data Model 1, $m_1$ columns are sampled randomly with replacement and any repeated columns are removed, $m_2$ rows are sampled randomly without replacement and the non-zero elements of $\bC$ are sampled independently from a zero-mean normal distribution.
If for fixed $c > 1$ and small $0 < \delta \ll 1/6$, $m_1$, $\beta$ and $q$ follow (\ref{suff thorem 1}), $\alpha$ is equal to (\ref{alpha eq}) and
\begin{eqnarray}
\begin{aligned}
 &  m_2 \ge \max \Bigg[ r \eta_u^2 \max \left( c_1 \log r , c_2 \log\left(\frac{3}{\delta}\right) \right) \: , \\
& \quad \quad \quad\quad \quad r+ q + 2 \log \frac{2}{\delta} + \sqrt{8 \: q \log \frac{2}{\delta} } \: , \\
& \quad \quad \quad\quad \quad r+ 1 +2 \log \frac{2 K}{\delta} + \sqrt{8 \: \log \frac{2 K}{\delta} }  \Bigg] \:,
\end{aligned}
\end{eqnarray}
where $c_1$ and $c_2$ are constant numbers, then Algorithm 1 with RRD yields the exact subspace and (\ref{outlier_detect 1}) identifies the non-zero columns of $\bC$ correctly with probability at least $1 - 6\delta$.
\label{Theorem 2}
\end{theorem}

\begin{remark}
In practice, the number of outliers is smaller than the number of inliers. Therefore, $\frac{K}{N_2^{'}} \leq 1$ (albeit this is not necessary for  Algorithm 1). Suppose that $C \mu_v^{'} r  \log \frac{4 r}{\delta} \ge 3 c^2 \frac{K}{N_2^{'}} \log \frac{2}{\delta}$, where $C$ is a constant number. According to (\ref{suff thorem 1}), it is almost sufficient to choose $\beta = 2$. Therefore, the sufficient number of randomly sampled columns
$
m_1 \ge 4 C \mu_v^{'} r \log \frac{4r}{\delta} \:,
$
i.e., $m_1$ scales linearly with $r \mu_v^{'}  \log 4r $.
The number of sampled outliers is $\calO(m_1  K/ N_2)$. Thus, the sufficient value for $m_2$ for  Algorithm 1 is $\calO( \max (r , \frac{K}{N_2^{'}} r \mu_v^{'} ) )$ with RED  and $\calO ( \max(r \eta_u^2 ,\frac{K}{N_2^{'}} r \mu_v^{'} ))$ with RRD.

%If we choose $c = 20$, then the maximum number of selected outlier columns can be bounded as
%\begin{eqnarray}
%q \leq C^{'}  \mu_v^{'} r  \log  \frac{2r}{\delta} \left( \frac{2K}{N_2^{'}} + 0.05 \right),
%\end{eqnarray}
%where $C^{'} = 4C$. Therefore, the order of sufficient value for $m_2$ is $\calO ()$

%can be rewritten as,
%\begin{eqnarray}
%m_2 \ge \frac{r \left( C^{'}  \mu_v^{'}   \log  \frac{8r}{\delta} \left( \frac{3K}{N_2^{'}} + 0.05 \right) +1 \right) \log 42\sqrt{2} + \log \frac{8}{\delta}}{f (1 / {2})} \:,
%\end{eqnarray}
%and for the second structure we can roughly say
%\begin{eqnarray}
%\begin{aligned}
% &  m_2 \ge \max \Bigg[ r \eta_u^2 \max \left( c_1 \log r , c_2 \log\left(\frac{3}{\delta}\right) \right) \: , \\
%& \quad \quad\quad \quad r+ q + 2 \log \frac{2}{\delta} + \sqrt{8 \: q \log \frac{2}{\delta} } \Bigg] \:,
%\end{aligned}
%\end{eqnarray}
\end{remark}

%\begin{remark}
%If we are just interested in locating the outlying columns, we can embed the whole data and use the learned columns subspace of $\bL_s^{\phi}$ to locate the outlying columns. This can be very useful for instance in settings where the data needs to be sent to a central processing unit, because we just need to send the embedded data matrix $\mathbf{\Phi} \bD$.
%\end{remark}

\subsection{Algorithm 2: randomized approach for the column-sparse outlier model}
\label{sec:algo2_sparse_outliers}
The table of Algorithm 2 details the randomized approach based on the column-sparsity of $\bC$ with both randomized designs. Algorithm 2 differs from Algorithm 1 in the subspace learning step since we do not assume that the outliers are linearly independent. Instead, subspace learning relies on the column-sparsity of $\bC$ and the convex algorithm (\ref{eq6}) is used in the subspace learning step. This implies a different requirement for the row compression step stated as follows.
 \begin{requirement}
The data sketching has to ensure that:\\
\textbf{1.} The rank of $\mathbf{\Phi} \bL$ is equal to $r$.\\
\textbf{2.} The non-zero columns of $\mathbf{\Phi} \bC_s$ do not lie in the column space of $\mathbf{\Phi} \bL$.
\end{requirement}

It is worth noting that the randomized approach substantially reduces the complexity of (\ref{eq6}). If (\ref{eq6}) is applied directly to $\bD$, the complexity would be $\calO(N_1 N_2 r)$ per iteration \cite{lamport10}. With the randomized approach, we show that the complexity of the subspace learning step is almost independent of the size of the data. The following theorems establish performance guarantees for Algorithm 2 with both randomized designs.

\begin{theorem}[Sufficient Condition-Algorithm 2 with RED]
Suppose $\bD$ follows Data model 1, the matrix $\mathbf{\Phi}$ is drawn from any distribution satisfying (\ref{eq12}), and the columns of $\bD_s$ are sampled randomly with replacement. If for small $0 < \delta \ll 1/3$,
\begin{eqnarray}
\begin{aligned}
& m_1 \ge \frac{N_2}{N_2^{'}}\zeta \: , \quad g \ge \frac{N_2^{'}}{N_2} \left( 1 + 6 r \mu_v (121/9) \right),\\
& \frac{K}{N_2^{'}} \leq \frac{g \frac{N_2}{N_2^{'}} - (1 + 6 r \mu_v (121/9))}{g (1+6 r \mu_v (121/9))} ,\\
&m_2 \ge \frac{(r+1) \log (42 \sqrt{2}) + \log K + \log \frac{2}{\delta}}{f (\frac{1}{{2}})},
\label{suff_cond    theorem 3}
\end{aligned}
\end{eqnarray}
where
\begin{eqnarray}
\zeta = \max \left( 3 g^2 \frac{K}{N_2^{'}} \log \frac{2}{\delta} \: , \frac{N_2^{'}}{N_2} 10\: r \mu_v  \log \frac{2r}{\delta} \right) \: ,
\label{g meghdar}
\end{eqnarray}
then Algorithm 2 with RED recovers the exact subspace and (\ref{outlier_detect 1}) correctly identifies the non-zero columns of $\bC$ with probability at least $1-3\delta$.
\label{Theorem 3}
\end{theorem}

\begin{theorem}[Sufficient Condition- Algorithm 2 with RRD]
Suppose $\bD$ follows Data model 1, the columns of $\bD_s$ are sampled randomly with replacement, and the rows are sampled randomly without replacement. In addition, it is assumed that the non-zero elements of $\bC$ are sampled independently from a zero-mean normal distribution.  If  for $0 < \delta \ll 1/4$, $m_1$, $g$ and $K / N_2^{'}$ follow (\ref{suff_cond    theorem 3}), $\zeta$ is  equal to (\ref{g meghdar}) and
\begin{eqnarray}
\begin{aligned}
&m_2 \ge \max \Bigg[ r \eta_u^2 \max \left( c_1 \log r , c_2\log\left(\frac{3 }{\delta}\right) \right) \: , \\
& \: r + 1 + 2 \log 2K /\delta + \sqrt{8 \log 2K /\delta}  \Bigg]
\label{suff theorem 4}
\end{aligned}
\end{eqnarray}
then Algorithm 2 with RRD recovers the exact subspace and (\ref{outlier_detect 1}) correctly identifies the non-zero columns of $\bC$ with probability at least $1-4\delta$.
\label{theorem 4}
\end{theorem}

\begin{remark}
If we choose
\begin{eqnarray}
 g = 2 \frac{N_2^{'}}{N_2} \big(1 + 6 r \mu_v (121/9)\big) \: ,
\end{eqnarray}
then the sufficient conditions (\ref{suff_cond    theorem 3}) can be rewritten as
\begin{eqnarray}
\begin{aligned}
& \frac{K}{N_2} \leq \frac{{N_2}/{2N_2^{'}}}{1+6 r \mu_v (121/9)}\\
&m_1 \hspace{-0.75mm}\ge \hspace{-0.75mm} \max \hspace{-0.75mm} \left(\hspace{-1mm}12 \frac{K}{N_2} \big(1 \hspace{-0.75mm}+ \hspace{-0.75mm} 6 r \mu_v (121/9)\big)^2 \hspace{-0.5mm} \hspace{-0.5mm} \log \frac{2}{\delta}\: , \: 10 r \mu_v  \log \hspace{-0.5mm}\frac{2r}{\delta} \hspace{-1mm}\right) \hspace{-0.5mm}\\
%&m_2 \ge \frac{(r+1) \log (42 \sqrt{2}) + \log q + \log \frac{2}{\delta}}{f (\frac{1}{{2}})} \: .
\end{aligned}
\end{eqnarray}
Thus, $m_1$ for Algorithm 2 is $\calO (\max(r\mu_v , \frac{K}{N_2} r^2 \mu_v^2))$. According to (\ref{suff_cond    theorem 3}) and (\ref{suff theorem 4}), the sufficient value for $m_2$ is roughly $\calO(r)$ with RED and $\calO(r \eta_v^2)$ with RRD. In addition, the permissible number of outliers scales linearly with $N_2$, i.e., not restricted to a sublinear sparsity regime.
\end{remark}

\section{Analysis of Algorithm 1}
\label{sec:analysis 1}
In this section, we provide a setp-by-step analysis of Algorithm 1. The proofs of the main theorems, the lemmas and the intermediate results are deferred to the appendix. First, we establish a sufficient condition on the number of sampled columns $m_1$ to guarantee that each inlier of $\bD_s$ lies in the span of the other inliers of $\bD_s$. Based on the number of sampled columns, we readily obtain an upper bound on the number of outlying columns in $\bD_s$. Then, we derive a sufficient condition for (\ref{eq11}) to satisfy Requirement 1.
\subsection{Random sampling from low rank matrices}
\label{sec sampling lw}
In the randomized approach, the column space of $\bL$ is learned from a small random subset of the columns of $\bD$. Therefore, we first have to ensure that the selected inliers span the column space of $\bL$. Initially, let's assume that $\bL^{'}\in \mathbb{R}^{N_1 \times N_2^{'}}$ is given.
%We derive a condition on the sufficient number of randomly sampled columns of $\bL^{'}$ to ensure that the selected columns can span its column space.
%Then, we use this result to establish a sufficient value for $m_1$ to guarantee that the inlier columns of $\bD_s$ can span the column space of $\bL$.
Suppose that $\bL^{'} = \bU^{'} \mathbf{\Sigma}^{'} (\bV^{'})^T$ is the compact SVD of $\bL^{'}$, where $\bU^{'} \in \mathbb{R}^{N_1 \times r}$, $\bV^{'} \in \mathbb{R}^{N_2^{'} \times r}$ and $\mathbf{\Sigma}^{'} \in \mathbb{R}^{r \times r}$.
The following lemma establishes a sufficient condition for a random subset of the columns of a low rank matrix to span its column space.
\begin{lemma}
Suppose $n_s$ columns are sampled uniformly at random with replacement from the matrix $\bL^{'}$ with rank $r$. If
\begin{eqnarray}
n_s \ge 10 \mu_v^{'} r  \log \frac{2 r}{\delta},
\label{eq22}
\end{eqnarray}
then the selected columns of the matrix $\bL^{'}$ span the column space of $\bL^{'}$ with probability at least $(1-\delta)$.
\label{lm4}
\end{lemma}
Hence, the column space of a low rank matrix $\bL^{'}$ can be captured from a small random subset of its columns when its row space is incoherent with the standard basis.

\subsection{Random column sampling from data matrix D}
Let $\alpha =20 \mu_v^{'} r  \log \frac{4 r}{\delta}$. Based on Lemma \ref{lm4}, the inliers in $\bD_s$ span the column space of $\bL$ and each inlier of $\bD_s$ lies in the span of the rest of the inliers of $\bD_s$ whp if the number of inliers in $\bD_s$ is at least $\alpha$.
%But, how many samples are needed to ensure that the number of inlier columns in $\bD_s$ is greater than or equal to $\alpha$?
Suppose we sample $m_1 = \beta \alpha { N_2}/{N_2^{'}}$ data columns randomly from $\bD$, where $\beta > 1$. The following lemma provides a sufficient condition on $\beta$ to ensure that the number of selected inliers exceeds $\alpha$.
% guarantee that the inlier columns of $\bD_s$ span the columns space of $\bL$ with high probability.
\begin{lemma}
Suppose that $ m_1 = \beta \alpha \frac{N_2}{N_2^{'}}$ columns of the given data matrix are sampled uniformly at random with replacement. If
\begin{eqnarray}
\beta \ge 2 + \frac{3}{\alpha} \log \frac{2}{\delta},
\label{eq34}
\end{eqnarray}
then the number of inlier columns of $\bD_s$ is greater than or equal to $\alpha$ with probability at least $(1 - \delta)$.
\label{lm6}
\end{lemma}
According to (\ref{eq34}), it is almost sufficient to choose $\beta = 2$. In addition, in most applications, $\frac{N_2}{N_2^{'}} < 2$. Therefore, if $4\alpha$ columns are sampled at random, the sampled columns will contain at least $\alpha$ randomly sampled inliers.

\subsection{Selected outlying columns}
The advantage of column sampling in the randomized approach is two-fold. First, complexity is substantially reduced since we only need to process a small subset of the data. Second, the number of outliers in $\bD_s$ is significantly smaller than the total number of outliers, which in turn relaxes the requirement on the spark of $\bC$ considerably. To clarify, robust PCA algorithms built on the linear independence assumption of the outlier columns as \cite{lamport8} require every subset of outliers with cardinality less than $(N_1 + 1)$ to be independent. In contrast, Algorithm 1 only requires independence for significantly smaller subsets of selected outliers.
%In the robust PCA algorithms which are built on the linear independence of outlier columns, usually it is assumed that any subset of outlying columns with cardinality less than $(N_1 + 1)$  are independent \cite{lamport8}. But, in Algorithm 1 we only need to assume that the selected outliers are independent which is substantially less than $N_1$ for high dimensional data. %
The following lemma establishes an upper-bound on the number of selected outliers.
\begin{lemma}
Suppose that $ m_1 = \beta \alpha \frac{N_2}{N_2^{'}}$ columns of the given data matrix are sampled uniformly at random with replacement. If
\begin{eqnarray}
\alpha \ge 3 c^2 \frac{K}{N_2^{'}} \log \frac{2}{\delta},
\label{eq37}
\end{eqnarray}
then the number of outliers selected is bounded from above by
\begin{eqnarray}
q = \alpha \left( \frac{\beta K}{N_2^{'}} + \frac{1}{c} \right)
\label{eq38}
\end{eqnarray}
with probability at least $(1 - \delta)$, where $c$ is any number greater than 1.
\label{lm7}
\end{lemma}

\subsection{Row compression}
In this section, we establish sufficient conditions on $m_2$ to satisfy Requirement 1. Suppose $\bD_s$ contains $k$ outlying columns. Thus, given Assumption 1, the rank of $\bD_s$ is equal to $r+k$. Requirement 1 is clearly satisfied if the rank of $\mathbf{\Phi} \bD_s$ is equal to the rank of $\bD_s$. The following lemmas provide sufficient conditions for $m_2$ with both randomized designs.
\begin{lemma}
Suppose $\bD_s$ contains at most $q$ outlying columns and assume that $\mathbf{\Phi}$ is an $m_2 \times N_1$ matrix satisfying the distributional JL property with
\begin{eqnarray}
m_2 \ge \frac{ (r + q) \log (42 \sqrt{2} ) + \log \frac{2}{\delta}}{f (\frac{1}{{2}})}.
\label{eq:  suffi m_2 1}
\end{eqnarray}
Then, the rank of $\mathbf{\Phi} \bD_s$ is equal to the rank of $\bD_s$ with probability at least $(1-\delta)$.
\label{lemma: embd 1}
\end{lemma}

\begin{lemma}
Suppose $\bD_s$ contains at most $q$ outlying columns, the rank of its low rank component $ \bL_s$ is equal to $r$, the non-zero elements of $\bC$ are sampled independently from a zero-mean normal distribution, and the rows of $\bD_s^{\phi}$ are $m_2$ randomly sampled (without replacement) rows of
$\bD_s$. If
\begin{eqnarray}
\begin{aligned}
&  m_2 \ge \max \Bigg[ r\eta_u^2 \max \left( c_1 \log r , c_2 \log\left(\frac{3}{\delta}\right) \right) \: , \\
& \quad \quad \quad \quad r+ q + 2 \log \frac{2}{\delta} + \sqrt{8 \: q \log \frac{2}{\delta} } \Bigg] \:,
\label{m_2 row 1}
\end{aligned}
\end{eqnarray}
where $c_1$ and $c_2$ are constant numbers,
then the rank of $\bD_s$ is equal to the rank of $\bD_s^{\phi}$ with probability at least $1 - 2\delta$.
\label{lemma row 1}
\end{lemma}

\section{Analysis of Algorithm 2}
\label{sec:analysis 2}
Similar to the analysis of Algorithm 1 in section \ref{sec:analysis 1}, we can make use of Lemma \ref{lm4} to derive a sufficient condition on $m_2$ to ensure that the rank of $\bL_s$ is equal to the rank of $\bL$. The number of selected outliers can also be bounded in a similar way.
The analysis in \cite{lamport10} established that (\ref{eq6}) yields exact outlier identification if $\bC$ is sufficiently column-sparse. If $\bC$ is sufficiently sparse, $\bC_s$ is also a column-sparse matrix whp.
Thus, we just need to ensure that $\bD_s^{\phi}$ is a representative data sketch with sufficient information. The following lemmas establish sufficient conditions on $m_2$ for the row compression step to satisfy Requirement 2 for both RED and RRD.

\begin{lemma}
Suppose $\bD_s$ contains at most $q$ outlying columns and assume that $\mathbf{\Phi}$ is an $m_2 \times N_1$ matrix satisfying the distributional JL property with
\begin{eqnarray}
m_2 \ge \frac{ (r + 1) \log (42 \sqrt{2}) + \log q + \log \frac{2}{\delta}}{f (\frac{1}{{2}})}.
\label{eq:  suffi m_2 2}
\end{eqnarray}
Then,  Requirement 2 is satisfied with probability at least $1 - \delta$.
\label{lemma: embd 2}
\end{lemma}

\begin{lemma}
Suppose the rank of $\bL_s$ is equal to $r$, $\bD_s$ contains at most $q$ columns, the non-zero elements of $\bC$ are sampled independently from a zero-mean normal distribution and the rows of $\bD_s^{\phi}$ are $m_2$ randomly sampled (without replacement) rows of $\bD_s$.
If
\begin{eqnarray}
\begin{aligned}
&  m_2 \ge \max \Bigg[ r\eta_u^2 \max \left( c_1 \log r , c_2 \log\left(\frac{3}{\delta}\right) \right) \: , \\
& \quad \quad  \quad r+ 1 +2 \log \frac{2 q}{\delta} + \sqrt{8 \: \log \frac{2 q}{\delta} } \Bigg] \:,
\label{m_2 row 2}
\end{aligned}
\end{eqnarray}
then Requirement 2 is satisfied with probability at least $1 - 2\delta$.
\label{lemma row 2}
\end{lemma}

\section{RED versus RRD and Complexity Analysis}
\begin{figure}[t!]
    \includegraphics[width=0.30\textwidth]{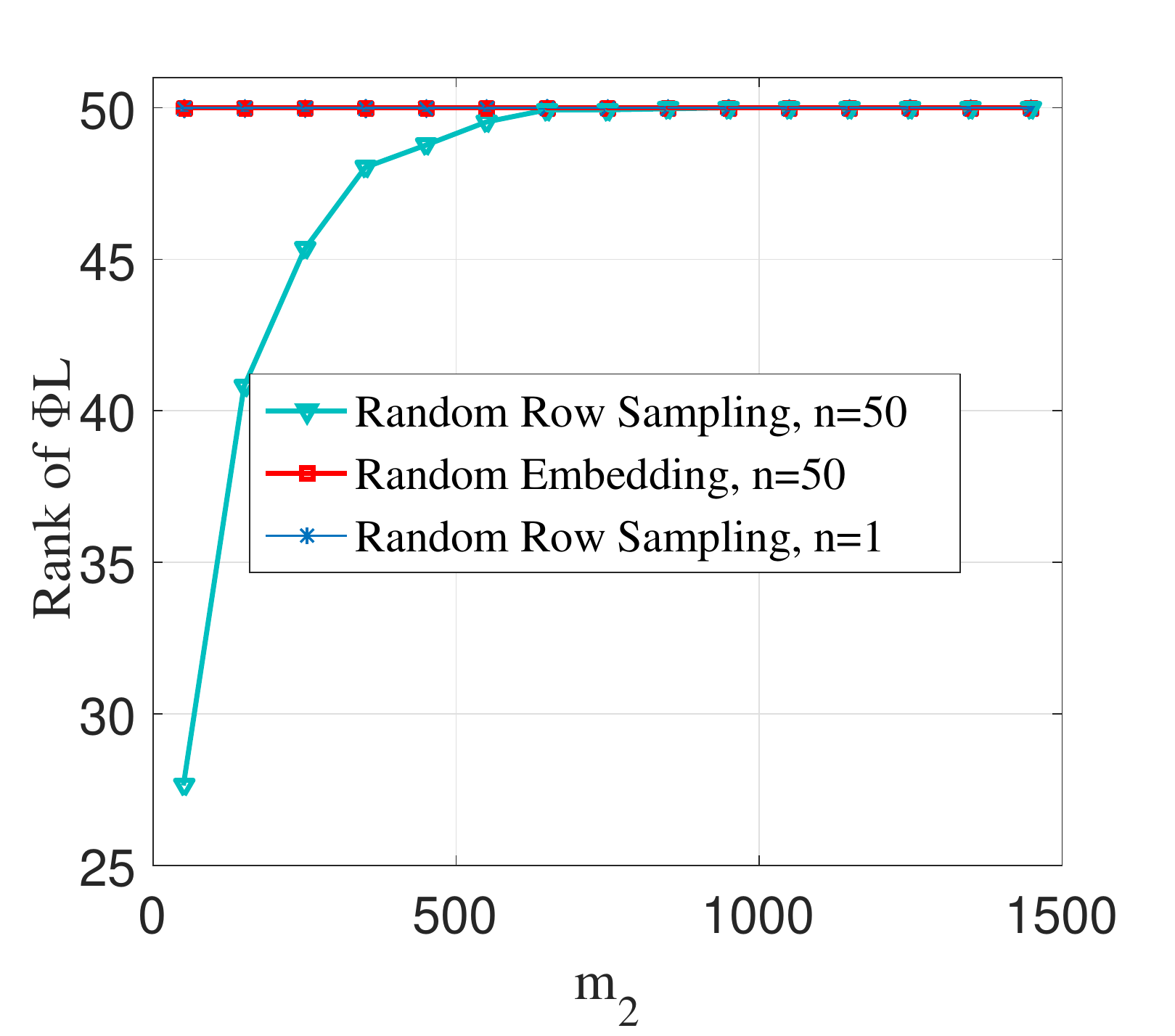}
    \centering
     \vspace{-0.2in}
    \caption{The rank of $\mathbf{\Phi}\bL$ versus $m_2$.}
    \label{fig: RED_RRD}
%   \vspace{-.5cm}
\end{figure}

%This section discusses the tradeoffs of both randomized designs.
 \textcolor{black}{ While the row compression step for RED has computational complexity $\calO (m_1 m_2 N_1)$ if we start data sketching with column sampling or  $\calO (m_2 N_1 N_2)$ if we start data sketching with row compression}, this step incurs no computational complexity in RRD.
%row compression step for RRD comes with no additional computational
%is that its row compression step does not impose any computation complexity. The computation complexity of row compression step with the RED is $\calO (m_2 N_1 N_2)$.
Hence, RRD may be more favorable for big data due to its reduced computational complexity. However, concerning sample complexity, random embedding is generally a more effective data sketching tool since the random projection matrix is not coherent with the data. To clarify, consider the extreme scenario where $r= 2$, $\bL \in \mathbb{R}^{2000 \times 2000}$ and only two rows of $\bL$ are non-zero. In this scenario, one needs to sample more or less the entire rows to ensure that the rank of $\bL^{\phi} = \mathbf{\Phi} \bL$ is equal to 2, i.e., $m_2$ has to be equal to 2000. In contrast, projecting the data into a random subspace with dimension equal to 2 is almost sufficient to ensure that the rank of $\bL^{\phi}$ is equal to 2 whp, i.e., $m_2 = 2$ is nearly sufficient. As another example, consider a
 %When the data points lie on a low dimensional subspace and the data points are not distributed uniformly in the subspace, we may not be able to capture the subspace with a small set of uniformly sampled data points. For instance, assume that
matrix $\bG \in \mathbb{R}^{2000 \times 3000}$ generated by concatenating the columns of matrices $\bG_i, i = 1, \ldots, n$, as
$
\bG = [ \bG_1 \: \bG_2 \: ... \: \bG_n ]  \:
$ and assume that $\bL = \bG^T$. For $1 \leq i \leq \frac{n}{2}$, $\bG_i = \bU_i \bQ_i\: ,$
where $\bU_i \in \mathbb{R}^{2000 \times \frac{r}{n}}$, $\bQ_i \in \mathbb{R}^{\frac{r}{n} \times \frac{100 r}{n}}$. For $n/2+1 \leq i \leq n$, $\bG_i = \bU_i \bQ_i\: ,$
where $\bU_i \in \mathbb{R}^{2000 \times \frac{r}{n}}$, $\bQ_i \in \mathbb{R}^{\frac{r}{n} \times \frac{20 r}{n}}$.
The elements of $\bU_i$ and $\bQ_i$ are sampled independently from a normal $\calN(0 , 1)$ distribution.
%and the elements of $\bU_i$  and $\bQ_i$ are sampled independently from a normal distribution $\calN(0 , 1)$.
The parameter $r$ is set equal to 50, thus, the rank of $\bL$ is equal to 50 whp. Accordingly, the rows of $\bL$ lie in a union of low-dimensional subspaces and if $n > 1$, the distribution of the rows of $\bL$ in the row space of $\bL$ will be highly non-uniform. Fig. \ref{fig: RED_RRD} shows the rank of $\mathbf{\Phi} \bL$ versus $m_2$.
When $n=1$, the rows of $\bL$ are distributed uniformly at random in the row space of $\bL$. Thus, $r$ rows sampled uniformly at random are enough to span the row space of $\bL$. But, when $n=50$, we need to sample almost 500 rows at random to span the row space. On the other hand, embedding the data into a random subspace with dimension 50 is almost sufficient to preserve the rank of $\bL$ even if $n = 50$.

\subsection{Computational complexity analysis}
The randomized approach consists of three steps: data sketching, subspace recovery and outlier detection. The data sketching step for RED has computational complexity $\calO (m_1 m_2 N_1)$ if data sketching starts with column sampling and $\calO (m_2 N_1 N_2)$ if it starts with row compression. Yet, this step has little impact on the actual run-time of the algorithms as it only involves a basic matrix multiplication operation for data embedding. Data sketching incurs no computational complexity in RRD. The complexity of subspace recovery is roughly $\calO (m_1^2 m_2)$ and $\calO (r m_1 m_2)$ for Algorithms 1 and 2, respectively. The outlier detection step (\ref{outlier_detect 1}) has complexity $\calO (m_2^2 N_2)$. As subspace learning and outlier detection (if intended) dominate the run-time of the algorithms, the randomized approach brings about substantial speedups in comparison to approaches that use the full-scale data. This is so given that the sufficient values for $m_1$ and $m_2$ are almost independent of the size of the data (cf. Section \ref{sec:prop_approach}), hence the randomized approach evades solving high-dimensional optimization problem. In contrast, solving (\ref{eq6}) for example has complexity $\calO(r N_1 N_2)$ per iteration. Table \ref{tabl 2} compares the run time of Algorithm 2 to the corresponding non-randomized approach with outlier detection. In this example, $r = 20$, $m_1 = 400$ and $m_2 = 100$. The randomized approach (even using RED) is remarkably faster than the non-randomized approach.
\begin{table}
\centering
\caption{Running time of randomized Algorithm 2 with outlier detection and the Algorithm in \cite{lamport10}. }
\vspace{-0.25cm}
\begin{tabular}{| c | c | c | c |}
\hline
$N_1 = N_2$& Algorithm 2 RED & Algorithm 2 RRD & \cite{lamport10}\\
 & + outlier detection & + outlier detection & \\
 \hline
 1000 & 0.5 s & 0.5 s & 30 s \\
 \hline
  5000 & 0.6 s & 0.6 s & 450 s \\
 \hline
  10000 & 1 s & 0.6 s & 2500 s \\
 \hline
  20000 & 2 s & 0.7 s & 12000 s \\
 \hline
\end{tabular}
\label{tabl 2}
\vspace{-.5cm}
\end{table}

\section{Noisy data}
In practice, noisy data can be modeled as
\begin{eqnarray}
\bD = \bL + \bC + \bN \:,
\end{eqnarray}
where $\bN$ is an additive noise component. %Both the convex decomposition problem (\ref{eq2}) and the representation learning problem (\ref{eq23}) were extended to noisy settings.
In \cite{lamport10}, it was shown that the optimal point of
\begin{eqnarray}
\begin{aligned}
& \underset{\hat{\bL},\hat{\bC}}{\min}
& & \lambda\|\hat{\bC}\|_{1,2}  + \|\hat{\bL} \|_* \\
& \text{subject to}
& & \big\| \hat{\bL} + \hat{\bC} - \bD \big\|_F \leq \epsilon_n \:, \\
\end{aligned}
\label{eq2noisy}
\end{eqnarray}
is equal to the optimal point of (\ref{eq6}) with an error proportional to the noise level.
The parameter $\epsilon_n$ has to be chosen based on the noise level. This modified version can be used in Algorithm 2 to account for the presence of noise.

Recall that Algorithm 1 is built on the idea that outliers of $\bD_s^{\phi}$ cannot be constructed from, or well-approximated by, linear combinations of the other columns of $\bD_s^{\phi}$. In the presence of noise, we further need to ensure that an outlier cannot be obtained from linear combinations of the columns of $\bN_s^{\phi} = \mathbf{\Phi} \bN \bS$. If an outlier lies in the span of the columns of $\bN_s^{\phi}$, the coefficients in the linear combinations of the columns of $\bN_s^{\phi}$ would have to be fairly large given that the columns of $\bN$ have small Euclidean norm.
%the columns of $\bN_s^{\phi}$ need big coefficients to constructs the outlying column since the columns of $\bN$ have small Euclidean norm.
Thus, to make Algorithm 1 robust to noise, we add a constraint to (\ref{eq13}) as follows
\begin{eqnarray}
\underset{\hat{\bz}}{\min} \quad \| \bd_{\phi s}^i -  \bQ_i^{\phi} \hat{\bz} \|_2 \quad \text{s.t.} \quad \| \hat{\bz} \|_p \leq \omega \:,
\end{eqnarray}
where $p \ge 1$ and $\omega$ is adjusted w.r.t. the noise level.

\section{Numerical Simulations}
%In the proposed approach, the given high-dimensional data is turned into small compressed data. First, a small random subset of the columns of the given data matrix is selected, then the selected columns are projected into a random low-dimensional subspace.  Therefore, the proposed method works with this small matrix instead of the given high-dimensional data.
In this section, we present some numerical experiments to study the requirements and performance of the randomized approach. The numerical results confirm that the sample complexity of the randomized methods is almost independent of the size of data. First, we investigate different scenarios using synthetic data. Then, the performance and requirements of the randomized algorithms are examined with real data.

%In this section, some numerical results are presented to demonstrate that the complexity of the proposed approach does not depend on the dimension of the given data.

%In addition, we study the effectiveness of the second method in dealing with linearly dependent outliers. In all simulations,

\subsection{Phase transition plots with synthetic data}
\begin{figure}[t!]
    \includegraphics[width=0.50\textwidth]{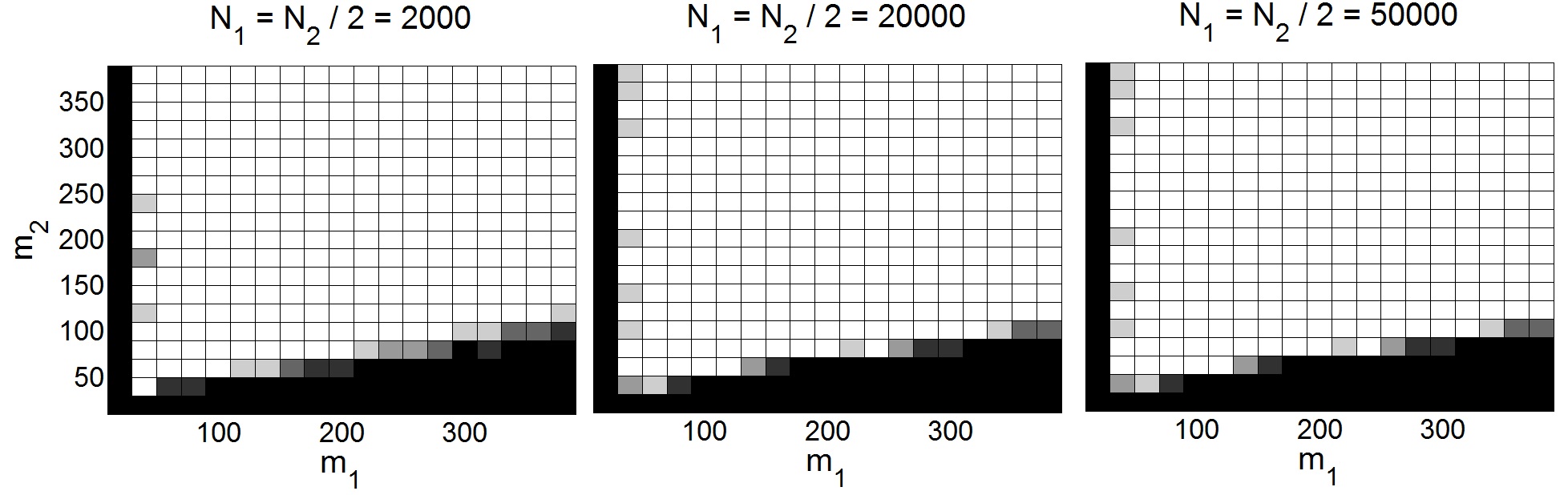}
    \centering
     \vspace{-0.3in}
    \caption{Phase transition plots of Algorithm 1 with RED ($r = 20$ , $\rho = 0.2)$. }
    \label{fig: 2000_50000}
\end{figure}

%A phase transition plot shows the performance of an algorithm for different values of $m_1$ and $m_2$.
In this section, The low rank matrix is generated as a product $\bL = \bU_r \bV_r^T$, where $\bU_r  \in \mathbb{R}^{N_1 \times r}$, $\bV_r  \in \mathbb{R}^{N_2 \times r}$. The elements of $\bU_r$ and $\bV_r$ are sampled independently from a standard normal $\mathcal{N} (0,1)$ distribution. The columns of $\bC$ are non-zero independently with probability $\rho$. Thus, the expected value of the number of outliers columns is $\rho N_2$. The non-zero entries of $\bC$ are sampled independently from $\calN (0 , 20^2)$. The phase transition plots show the probability of correct subspace recovery for the pairs of $(m_1 , m_2)$. White designates exact subspace recovery and black indicates incorrect recovery. In all experiments presented in this section the data is a $2000 \times 4000$ matrix except for the simulation in Fig. \ref{fig: 2000_50000}.

\begin{figure}[t!]
    \includegraphics[width=0.50\textwidth]{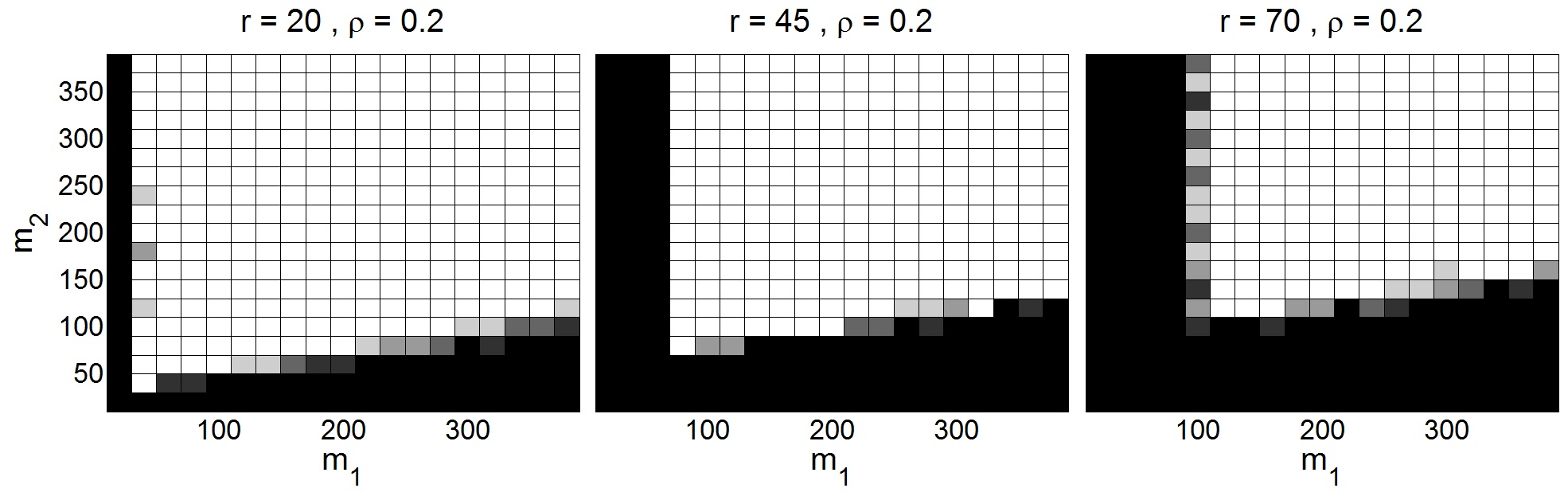}
    \centering
     \vspace{-0.3in}
    \caption{Phase transition plots of Algorithm 1 with RED}
    \label{fig: frs_alg1_r}
\end{figure}

\begin{figure}[t!]
    \includegraphics[width=0.50\textwidth]{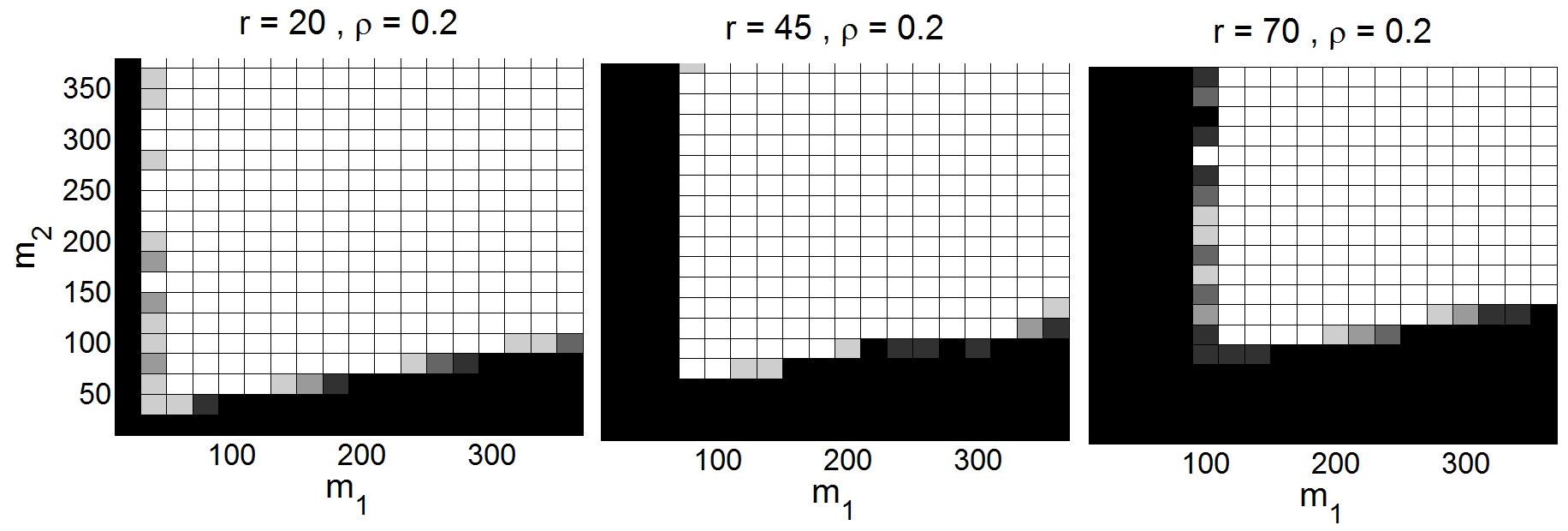}
    \centering
     \vspace{-0.3in}
    \caption{Phase transition plots of Algorithm 1 with RRD}
    \label{fig: srs_alg1_r}
\end{figure}

Fig. \ref{fig: frs_alg1_r} shows the phase transition of Algorithm 1 with RED for different values of $r$. When $r$ is increased, the required values of $m_1$ and $m_2$ increase as we need more samples to ensure that the selected columns span the column space of $\bL$, as well as a higher dimension for the embedding subspace given that the column space of $\bL$ has a higher dimension.
%In addition, when the rank is increased, the dimension of the column space of $\bL$ increases, implying that we need a random subspace with higher dimension. Hence, the required value of $m_2$ also increases.
Fig. \ref{fig: srs_alg1_r} shows a similar plot with RRD. Since in this section the columns/rows of $\bL$ are distributed uniformly at random in the column/row-space of $\bL$, RED and RRD yield a similar performance. As such, for the remaining scenarios in this section we only provide phase transitions with RED (RRD yields the same performance).
%Since in this section the rows of $\bL$ are distributed randomly, we provide the phase transitions only with the RED  for the subsequent scenarios (the RRD yields the same performance).

\begin{figure}[t!]
    \includegraphics[width=0.50\textwidth]{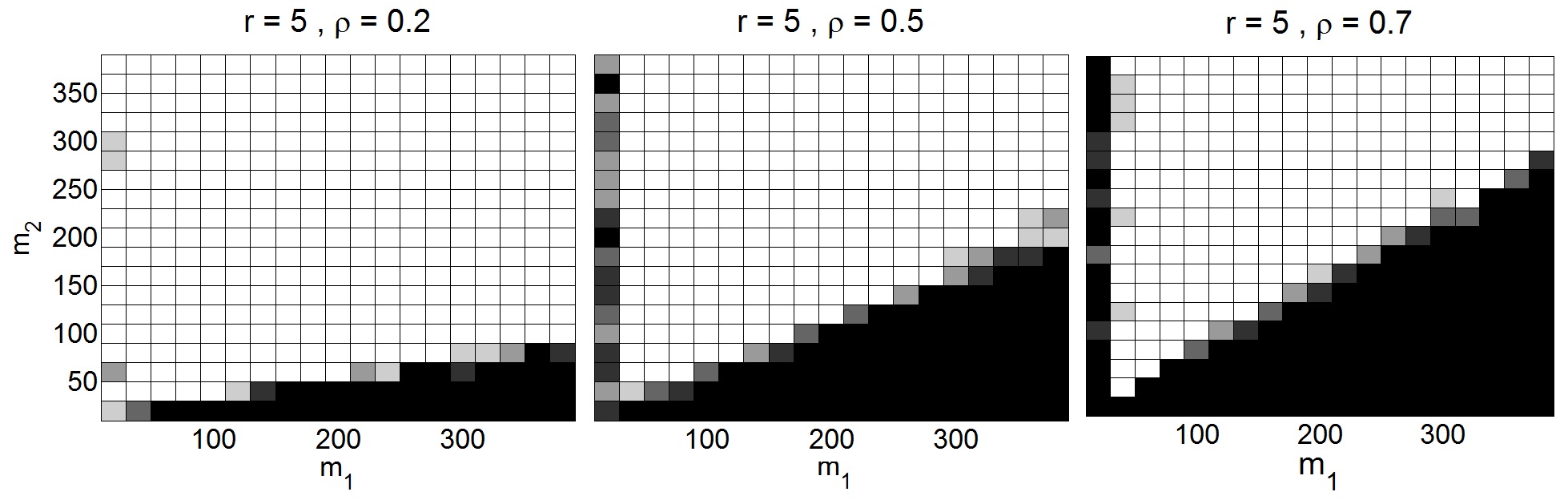}
    \centering
     \vspace{-0.3in}
    \caption{Phase transition plots of Algorithm 1 with RED}
    \label{fig: frs_alg1_rho}
\end{figure}

Fig. \ref{fig: frs_alg1_rho} illustrates the phase transition for Algorithm 1 with RED  for different values of $\rho$. Increasing $\rho$ has only minimal effect on $m_1$ (which is almost around 25) because the required number of sampled columns depends linearly on $r \frac{1}{(1- \rho)}$. Therefore, when $\rho$ is increased from 0.2 to 0.7, $r \frac{1}{(1- \rho)}$ increases from $1.25 r$ to $3.3 r$. It is interesting to observe that when the number of sampled columns is increased, the required $m_2$ also increases. This is due to the fact that the number of sampled outlier columns increases as we sample more columns. Subsequently, the selected outliers span a subspace with a higher dimension, wherefore we need a random subspace with higher dimension for embedding the sampled columns because to ensure that the rank of $\mathbf{\Phi} \bD_s$ is equal to the rank $\bD_s$ in Algorithm 1.

\begin{figure}[t!]
    \includegraphics[width=0.50\textwidth]{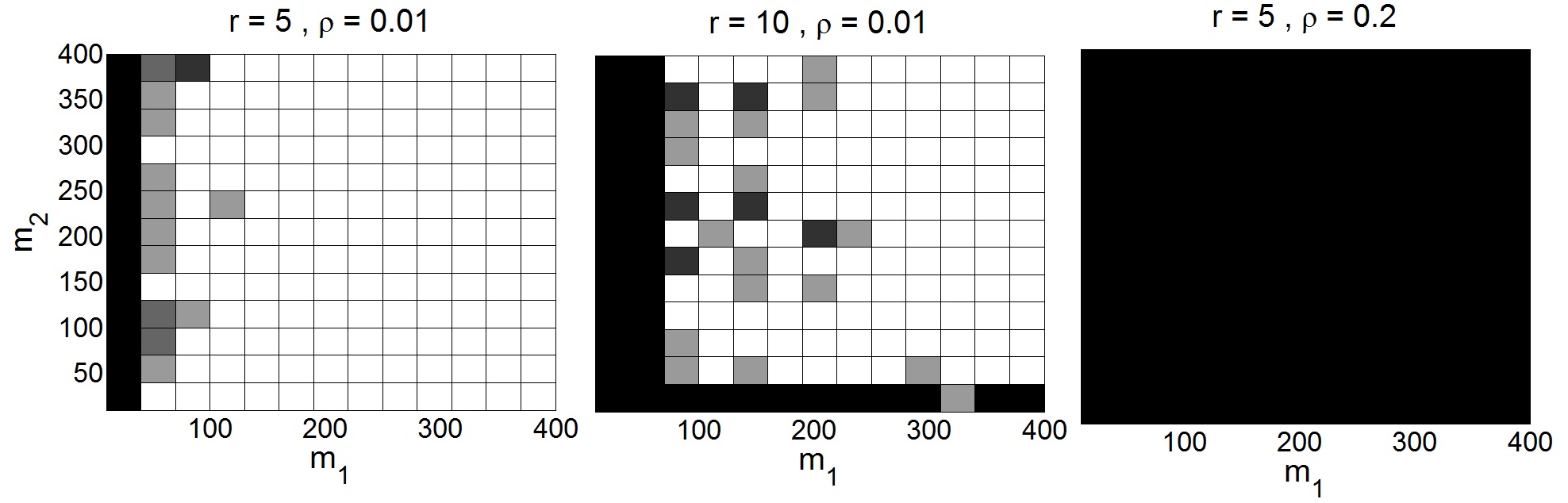}
    \centering
     \vspace{-0.3in}
    \caption{Phase transition plots of Algorithm 2 with RED.}
    \label{fig: convex_phase}
\end{figure}

The phase transition plots for Algorithm 2 with RED are shown in Fig. \ref{fig: convex_phase} for different values of $\rho$ and $r$. In the left plot, $r = 5$ and $\rho = 0.01$. With $m_1 > 100$ and $m_2 \ge 50$, the algorithm yield correct output whp. In the middle plot, the rank is increased to 10. Thus, the required values for $m_1$ and $m_2$ increase. In the right plot, $\rho = 0.2$ and Algorithm 2 cannot yield correct subspace recovery since (\ref{eq6}) requires $\bC$ to be column-sparse (roughly requiring $\rho \leq 0.05$).

Fig. \ref{fig: 2000_50000}
shows the phase transition of Algorithm 1 with RED for data matrices with different dimensions. Although the size of the data is increased from $2000 \times 4000$ to $(5 \times 10^4) \times 10^5$, the required values for $m_1$ and $m_2$ remain unchanged confirming our analysis, which revealed that the sample complexity of the proposed approach is almost independent of the size of the data. In this simulation, since the columns/rows are distributed randomly, the column space and row space of $\bL$ have small incoherence parameters \cite{lamport49}. Thus the factors dominating the sample complexity are $r$ and $\rho$.

\begin{figure}[t!]
    \includegraphics[width=0.40\textwidth]{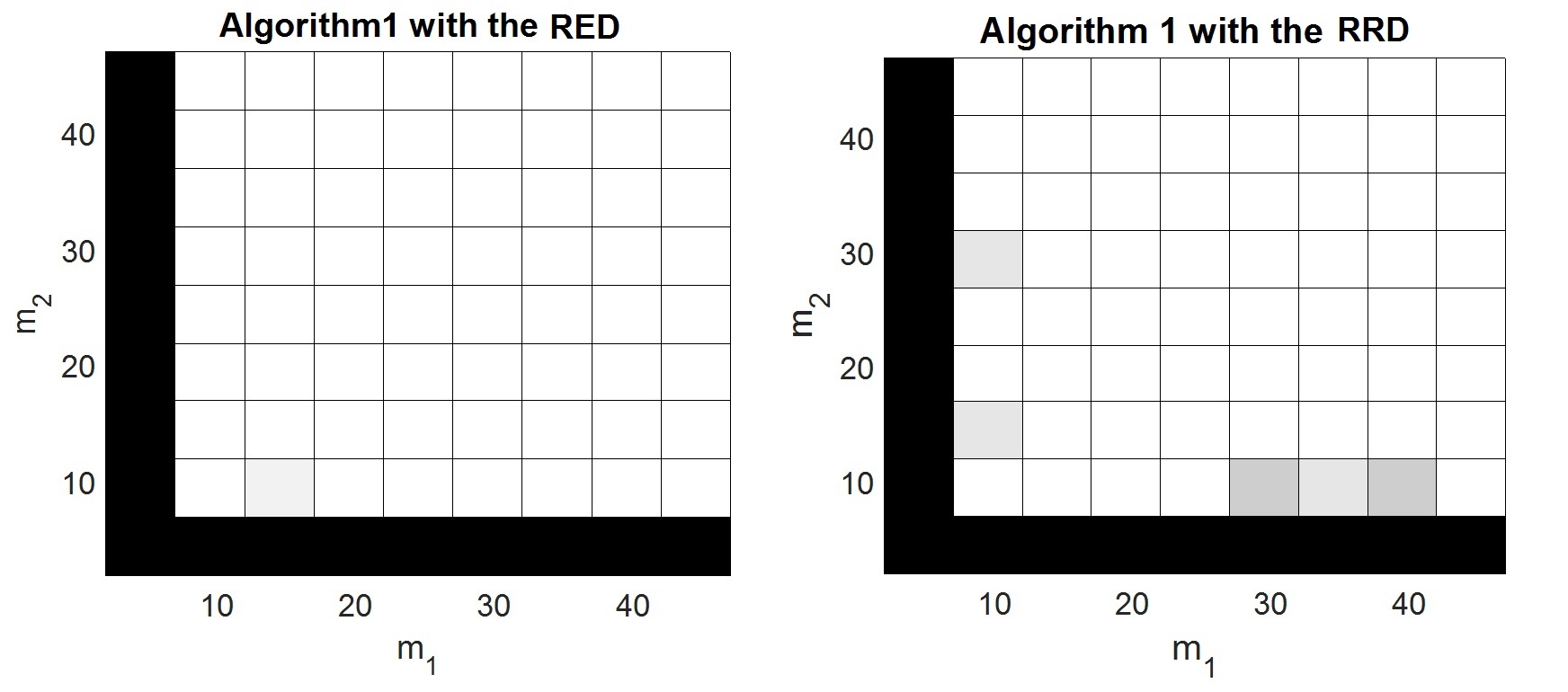}
    \centering
     \vspace{-0.1in}
    \caption{Phase transition plots of Algorithm 1 with both RED  and RRD applied to motion tracking data. }
    \label{fig: motion}
\end{figure}

\begin{figure}[t!]
    \includegraphics[width=0.40\textwidth]{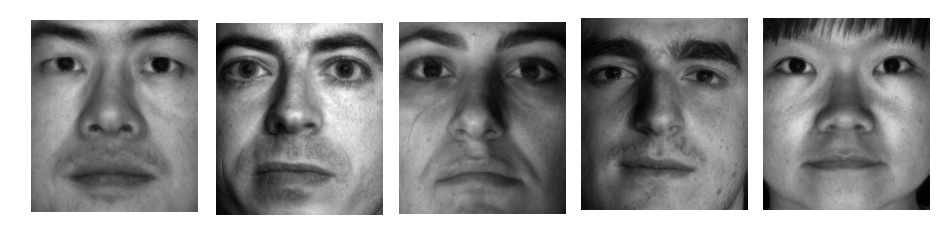}
    \centering
     \vspace{-0.1in}
    \caption{A set of random examples of the faces in Yale database. }
    \label{fig: yale example}
\end{figure}

\begin{figure}[t!]
    \includegraphics[width=0.40\textwidth]{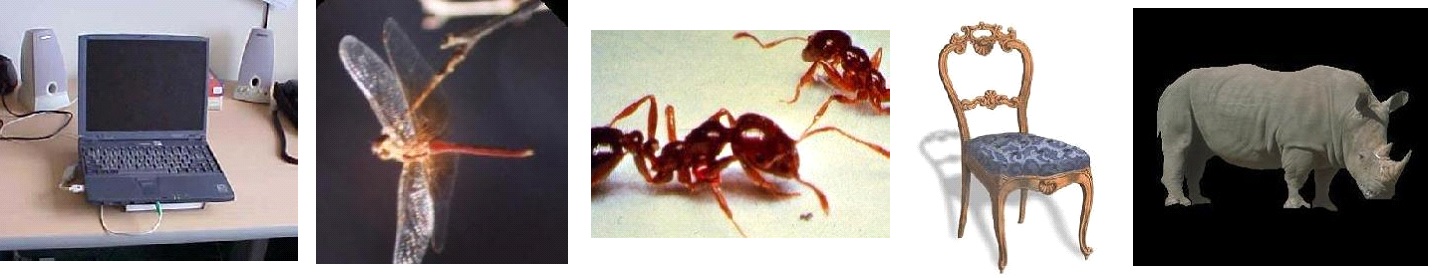}
    \centering
     \vspace{-0.1in}
    \caption{Random examples of the images in Caltech101 database. }
    \label{fig: caltech ex}
\end{figure}

\begin{figure}[t!]
    \includegraphics[width=0.40\textwidth]{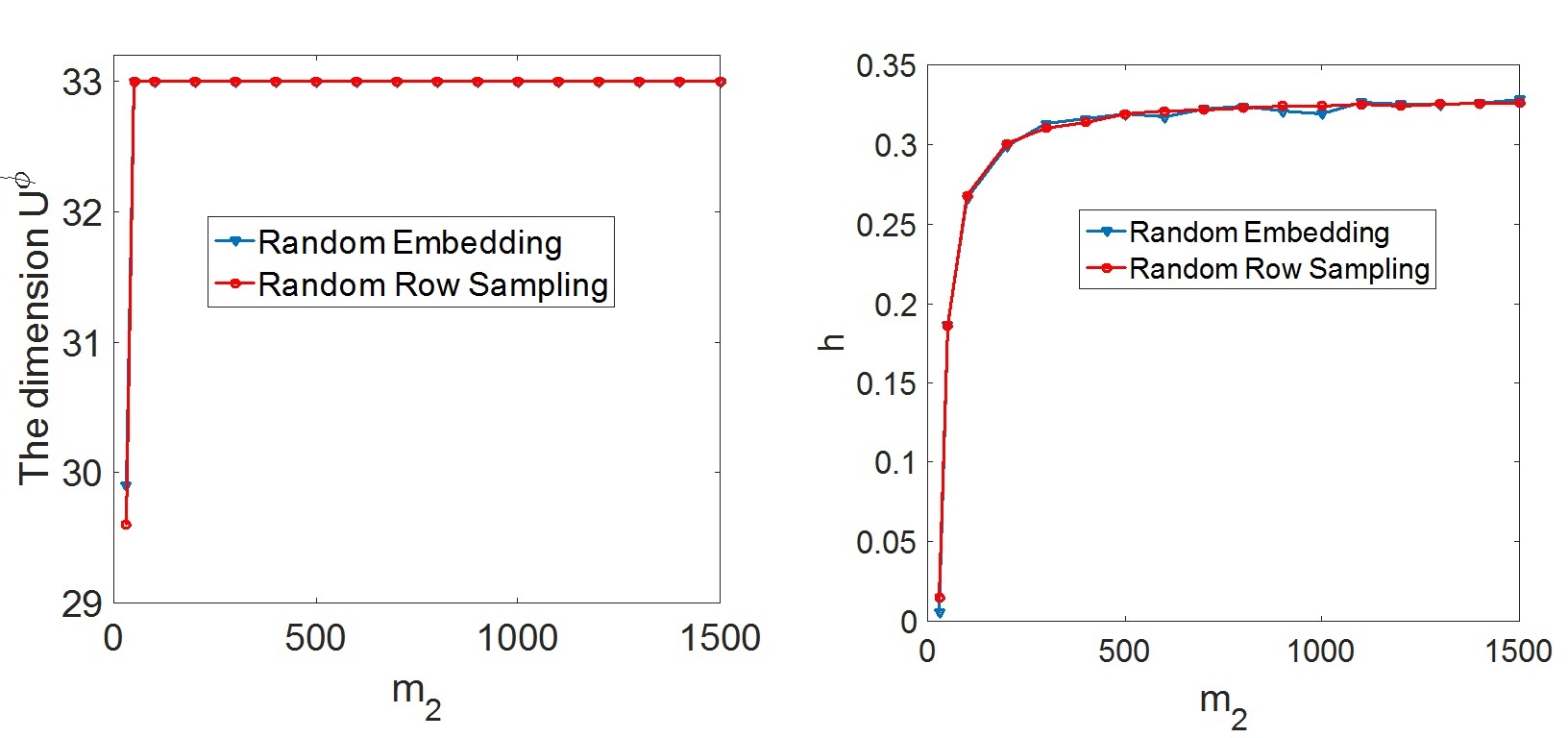}
    \centering
     \vspace{-0.1in}
    \caption{The dimension of $\bU^{\phi}$ and $h$ versus the value of $m_2$. }
    \label{fig: dimension of face}
\end{figure}

\subsection{Phase transition with real data}
In this section, we study the requirements of the randomized approach with real data for motion tracking and segmentation. The  data  is  generated  by  extracting  and  tracking  a  set  of
points  throughout the frames \cite{tron2007benchmark}. The data is a low rank matrix, and the motion data points lie in a union of low-dimensional subspaces. We use one of the scenarios in Hopkins155 \cite{tron2007benchmark}. This data matrix is $62 \times 464$ and its rank is roughly equal to 4. We add 50 outlying data points. Thus, the final data is $62 \times 512$. Fig. \ref{fig: motion} is the phase transition of Algorithm 1 with RED and RRD showing the probability of correct outlier identification. When $m_1$ and $m_2$ are greater than 10, the algorithm yields exact outlier detection whp.

\subsection{Sufficient values for $m_2$ with face images}
Vectorized images are high dimensional data vectors. Thus, if they construct low dimensional subspaces, substantial reductions in computational complexity and memory requirements can be achieved through the row compression operation of the randomized approach. In this experiment, we use the face images in the Extended Yale Face Database B \cite{lee2005acquiring} as inlier data points. Fig. \ref{fig: yale example} displays a random subset of these faces. This database consists of face images from 38 human subjects, and the images of each subject lie in a low-dimensional subspace \cite{lee2005acquiring}. According to our investigations, the dimension of the face images (38 faces) is roughly equal to 33. We randomly sample 350 images of the Caltech101 database \cite{fei2007learning} as outlying data points. Fig. \ref{fig: caltech ex} displays a randomly chosen set of the images in the Caltech101 database.
 Define $\bU$ as a basis for the subspace of the faces, $\bU^{\phi} = \mathbf{\Phi}\bU$ and ${\bC^{\phi}}^{\perp}$ as
\begin{eqnarray}
{\bC^{\phi}}^{\perp} = \bigg( \bI  - \bU^{\phi} ( {\bU^{\phi}}^T \bU^{\phi} )^{-1} {\bU^{\phi}}^T \bigg) \bC^{\phi} \:.
\end{eqnarray}
In addition, define
\begin{eqnarray}
h = \| {\bC^{\phi}}^{\perp} \|_F / \| \bC^{\phi} \|_F \:.
\end{eqnarray}
Thus, by measuring the dimension of the span of $\bU^{\phi}$ and the value of $h$, we can observe if the row compression operation preserves the essential information since the dimension of $\bU^{\phi}$ is the rank of of the low rank component and $h$ is proportional to the norm of the components of the outlying data points which do not lie in the column space of the low rank component. Fig. \ref{fig: dimension of face} shows the dimension of $\bU^{\phi}$ and the values of $h$ versus $m_2$ for both random embedding and random row sampling. Although the dimension of the data vectors is 32256, it is shown that 300 random linear measurements of the data vectors are nearly sufficient to preserve the rank of $\bL$ and the outlying component of $\bC$.

%\subsection{Random Embedding versus Random Sampling}
%Although the complexity of row compression step is
\section{Appendix}
\noindent\textbf{Proof of lemma \ref{lm4}}\\
The matrix of sampled columns can be represented as
\begin{eqnarray}
\bL_s^{'} = \bL^{'} \bS^{'}
\label{eq23}
\end{eqnarray}
where $\bS^{'} \in \mathbb{R}^{N_2^{'} \times n_s}$ selects the columns to sample. Using the SVD of $\bL^{'}$, (\ref{eq23}) can be rewritten as
\begin{eqnarray}
\bL_s^{'} = \bU^{'} \mathbf{\Sigma}^{'} (\bV^{'})^T \bS^{'} \quad.
\label{eq24}
\end{eqnarray}
Therefore, if the matrix $ (\bV^{'})^T \bS^{'}$ is full rank, the selected columns of $\bL^{'}$ span its column space.

Define $\bs_i^{'}$ as the $i^{\text{th}}$ column of $\bS^{'}$. The vector $\bs_i^{'}$ can be any of the vectors of the standard basis with equal probability since we are using random sampling with replacement. Therefore,
\begin{eqnarray}
\mathbb{E} \left[ (\bV^{'})^T \bs_i^{'} (\bs_i^{'})^T \bV^{'} - \frac{1}{N_2^{'}} \bI \right] = 0 .
\label{es1}
\end{eqnarray}
The matrix $ (\bV^{'})^T \bS^{'} (\bS^{'})^T \bV^{'} $ can be written as
\begin{eqnarray}
 (\bV^{'})^T \bS^{'} (\bS^{'})^T \bV^{'} = \sum_{i=1}^{n_s}  (\bV^{'})^T \bs_i^{'} (\bs_i^{'})^T \bV^{'}.
\label{es2}
\end{eqnarray}
If $(\bV^{'})^T \bS^{'} (\bS^{'})^T \bV^{'}$ is a full rank matrix, then $(\bV^{'})^T \bS^{'}$ is also full rank. In addition, if we can show that
\begin{eqnarray}
\Big \|  (\bV^{'})^T \bS^{'} (\bS^{'})^T \bV^{'}  - \frac{n_s}{N_2^{'}} \bI \Big \|
\label{es3}
\end{eqnarray}
is sufficiently small, we can conclude that $ (\bV^{'})^T \bS^{'} (\bS^{'})^T \bV^{'} $ is full rank. According to (\ref{es1}) and (\ref{es2}), the matrix
\begin{eqnarray}
 (\bV^{'})^T \bS^{'} (\bS^{'})^T \bV^{'} - \frac{n_s}{N_2^{'}} \bI
\label{es4}
\end{eqnarray}
is a sum of $n_s$ independent zero-mean random matrices. Thus, we use the non-commutative Bernstein Inequality \cite{lamport11} to bound the spectral norm of (\ref{es4}).

\begin{lemma}[\text{Non-commutative Bernstein Inequality \cite{lamport11}}]
Let $\bX_1 , \bX_2 , ... , \bX_L $ be independent zero-mean random matrices of dimension $d_1 \times d_2$. Suppose $\rho_k^2 = \max \{\| \mathbb{E} [ \bX_k \bX_k^T ] \| , \| \mathbb{E} [ \bX_k^T \bX_k ] \| \} $ and $\| \bX_k \| \leq M$ almost surely for all k. Then for any $\tau > 0$
\begin{eqnarray}
\begin{aligned}
&\mathbb{P} \left[ \Bigg \| \sum_{k=1}^{L} \bX_k \Bigg  \| > \tau \right] \leq \\
&(d_1 + d_2) \exp \left( \frac{-\tau^2 / 2}{\sum_{k=1}^L \rho_k^2  + M\tau/3} \right) .
\end{aligned}
\label{eq28}
\end{eqnarray}
\label{lm5}
\end{lemma}
 In our problem, $\bX_i = (\bV^{'})^T \bs_i^{'} (\bs_i^{'})^T \bV^{'} - \frac{1}{N_2^{'}} \bI $. If the matrices $\bA$ and $\bB$ are positive definite, then $\| \bA - \bB \| \leq \max \{ \| \bA \|, \| \bB \| \} $. Thus, we can derive $M$ as follows
\begin{eqnarray}
\begin{aligned}
\Big \| (\bV^{'})^T &\bs_i^{'} (\bs_i^{'})^T \bV^{'} - \frac{1}{N_2^{'}} \bI \Big \| \\
&\leq  \max \{ \Big  \| (\bV^{'})^T \bs_i^{'} (\bs_i^{'})^T \bV^{'} \Big  \| , \Big \| \frac{1}{N_2^{'}} \bI \Big \|  \} \leq \frac{r \mu_v^{'}}{N_2^{'}}.
\label{eq29}
\end{aligned}
\end{eqnarray}
 We also have
\begin{eqnarray}
\nonumber
\begin{aligned}
&\Bigg \| \mathbb{E} \left[ \left(  (\bV^{'})^T \bs_i^{'} (\bs_i^{'})^T \bV^{'} - \frac{1}{N_2^{'}} \bI\right) \left(  (\bV^{'})^T \bs_i^{'} (\bs_i^{'})^T \bV^{'} - \frac{1}{N_2^{'}} \bI \right) \right] \Bigg \| \\
& = \Bigg \| \mathbb{E} \left[ (\bV^{'})^T \bs_i^{'} (\bs_i^{'})^T \bV^{'} (\bV^{'})^T \bs_i^{'} (\bs_i^{'})^T \bV^{'}  - \frac{1}{(N_2^{'})^2}  \bI \right] \Bigg \| \\
& \leq \max \Bigg \{ \Big \| \mathbb{E} \left[ (\bV^{'})^T \bs_i^{'} (\bs_i^{'})^T \bV^{'} (\bV^{'})^T \bs_i^{'} (\bs_i^{'})^T \bV^{'}  \right] \Big \| , \frac{1}{(N_2^{'})^2} \Bigg \} \\
& \leq \max \Bigg \{ \frac{r \mu_v^{'}}{N_2^{'}}  \Big \| \mathbb{E} \left[ (\bV^{'})^T \bs_i^{'} (\bs_i^{'})^T \bV^{'}  \right] \Big \| , \frac{1}{(N_2^{'})^2} \Bigg \} \leq  \frac{r \mu_v^{'}}{(N_2^{'})^2} \quad.
\end{aligned}
\label{eq30}
\end{eqnarray}
Therefore, according to Lemma \ref{lm5} if we set
\begin{eqnarray}
n_s \ge \frac{28}{3} r \mu_v^{'} \log \frac{2 r}{\delta}
\label{eq31}
\end{eqnarray}
then,
\begin{eqnarray}
\mathbb{P} \left[ \frac{N_2^{'}}{n_s} \Bigg \|  (\bV^{'})^T \bS^{'} (\bS^{'})^T \bV^{'}  - \frac{n_s}{N_2^{'}} \bI  \Bigg  \| > \frac{1}{2} \right] \leq \delta \quad.
\label{eq32}
\end{eqnarray}
If $\sigma_1$ and $\sigma_r$ denote the largest and smallest singular values of $(\bS^{'})^T \bV^{'}$, respectively, then
\begin{eqnarray}
\frac{n_s}{2 N_2^{'}} \leq \sigma_1^2 \leq \sigma_r^2 \leq \frac{3 n_s}{2 N_2^{'}} \: .
\label{eq33}
\end{eqnarray}
 Accordingly, the matrix $(\bS^{'})^T \bV^{'}$ is a full rank matrix with probability at least $1- \delta$.
\smallbreak

\noindent\textbf{Proof of Lemma \ref{lm6}}\\
Since we use random sampling with replacement, the number of inliers in the selected columns follows a Binomial distribution. Suppose $n_i$ is the number of sampled inlier columns. Then, $n_i$ is a Binomial random variable with $m_1$ independent experiments, each with success probability $\frac{N_2^{'}}{N_2}$. Therefore, using Chernoff bound for Binomial distributions \cite{lamport42}, we have
\begin{align}
%\begin{aligned}
\mathbb{P} \left( \alpha \leq n_i  \leq  \alpha(2\beta -1)  \right)   \ge
 1 - 2 \exp \left( - \frac{\alpha^2 (\beta - 1)^2}{3 \alpha \beta}\right).
%\end{aligned}
\label{eq35}
\end{align}
Thus, if
%\begin{eqnarray}
$\beta \ge 2 + \frac{3}{\alpha} \log \frac{2}{\delta}$,
%\label{eq36}
%\end{eqnarray}
 the RHS of (\ref{eq35}) is lower-bounded by $(1 - \delta)$.
\smallbreak

\noindent\textbf{Proof of Lemma \ref{lm7}}\\
Since we use random sampling with replacement, the number of outliers $n_o$ in the matrix $\bD_s$ follows a Binomial distribution %Suppose $o_i$ is the number of sampled outlier columns.
with $m_1$ independent experiments, each having success probability $\frac{K}{N_2}$. Using Chernoff bound we have that
\begin{align}
%\begin{aligned}
&\mathbb{P} \left( \alpha \left( \frac{\beta K}{N_2^{'}} - \frac{1}{c} \right) \leq n_o  \leq  \alpha \left( \frac{\beta K}{N_2^{'}} + \frac{1}{c} \right)  \right) \\
& \ge 1 - 2 \exp \left( - \frac{\alpha^2}{3 c^2 \alpha \beta \frac{K}{N_2^{'}}} \right) \:.
%\end{aligned}
\label{eq39}
\end{align}
Therefore, the RHS of (\ref{eq39}) is greater than $1 - \delta$ if (\ref{eq37}) is satisfied.
\\
%\bibliographystyle{IEEEtran}

\noindent\textbf{Proof of Lemma \ref{lemma: embd 1}}\\
To prove Lemma \ref{lemma: embd 1} and Lemma \ref{lemma: embd 2}, we make use of the following result from  \cite{lamport39} and \cite{lamport40}.
\begin{lemma}
Let $\cup_{i=1}^{n} \mathcal{L}_i$ denote a union of $n$ linear subspaces in $\mathbb{R}^{N_1}$, each of dimension at most $d$. For fixed $\delta \in (0,1)$ and $\epsilon \in (0,1)$, suppose $\mathbf{\Phi}$ is an $m_2 \times N_1$ matrix satisfying the distributional JL property with
\begin{eqnarray}
m_2 \ge \frac{d \log (42/\epsilon) + \log n + \log \frac{2}{\delta}}{f (\frac{\epsilon}{\sqrt{2}})}.
\label{eq48}
\end{eqnarray}
Then,
\begin{eqnarray}
(1 - \epsilon) \| \bv \|_2^2 \leq  \| \mathbf{\Phi}\bv \|_2^2 \leq (1 + \epsilon) \| \bv \|_2^2
\label{eq49}
\end{eqnarray}
holds simultaneously for all $\bv \in \cup_{i=1}^{K} \mathcal{L}_i$ with probability at least $(1-\delta)$.
\label{lm10}
\end{lemma}

According to Lemma \ref{lm10}, if $m_2$ satisfies (\ref{eq:  suffi m_2 1}), then (\ref{eq49}) holds with $\epsilon = 1/\sqrt{2}$ for all the vectors in the column space of $\bD_s$ with probability at least $1-  \delta$. If (\ref{eq49}) holds for all $\bv \in \spn(\bD_s)$, then it is straightforward to show that the rank of $\mathbf{\Phi \bD_s}$ is equal to the rank of $\bD_s$.
\smallbreak

\noindent\textbf{Proof of Lemma \ref{lemma: embd 2}}\\
Suppose $\bD_s$ contains $k$ outlying data points. Assume $\{ \cup \mathcal{T}_i \}_{i=1}^{k}$ represents a union of $k$ linear subspaces in $\mathbb{R}^{N_1}$, where each subspace is spanned by $\{ \bU, \bC_i \}$ and $\bC_i$ is the $i^{\text{th}}$ non-zero column of $\bC_s$. According to Data model 1, the subspace $\mathcal{T}_i$ is an $(r+1)$-dimensional subspace since $\bC_i$  does not lie in the column space of $\bL$. Suppose $\mathbf{\Phi}$ is a stable embedding of the union of subspaces $\{ \cup \mathcal{T}_i \}_{i=1}^{k}$. Then, the dimension of the subspaces $\{ \mathcal{T}_i \}_{i=1}^{k}$ is not changed during the embedding operation. Accordingly, the columns of $ \mathbf{\Phi} \bC_s$ do not lie in the column space of $ \mathbf{\Phi} \bL$. Note that $q \ge k$. Thus, according to Lemma \ref{lm10}, if
\begin{eqnarray}
m_2 \ge \frac{(r+1) \log (42 \sqrt{2}) + \log q + \log \frac{2}{\delta}}{f (\frac{1}{{2}})}.
\end{eqnarray}
then the rank of $\mathbf{\Phi} \bL$ is equal to the rank of $\bL$ and the non-zero columns of $\mathbf{\Phi} \bC_s$ do not lie in the column space of $\mathbf{\Phi} \bL$, with probability at least $1 - \delta$.
\smallbreak
\noindent\textbf{Proof of Lemma \ref{lemma row 1}}\\
Since the rank of $\bL_s$ is equal to $r$, $\bL$ and $\bL_s$ have the same column space.
Suppose $\bC_s$ contains $k$ outlying columns.
We break this proof into two steps. In the first step, it is shown that the rank of $\bL_s$ is equal to the rank of $\mathbf{\Phi} \bL_s$ whp.
Define ${\bU_s^{\phi}}^\perp$ as an orthonormal basis for the complement of the column space of $\bL_s^{\phi} = \mathbf{\Phi} \bL_s$. If the rank of $\bL_s^{\phi}$ is equal to $r$, then ${\bU_s^{\phi}}^\perp \in \mathbb{R}^{m_2 \times (m_2 - r)}$.
 In the second step, it is proven that the rank of
\begin{eqnarray}
({\bU_s^{\phi}}^\perp)^T\bC_s^{\phi}
\label{eq proj}
\end{eqnarray}
is equal to $k$ whp. The matrix (\ref{eq proj}) is the projection of the columns of $\bC_s^{\phi}$ onto the complement of the column space of $\bL_s^{\phi}$. Lemma \ref{lemma row 1} follows if these two requirements are satisfied.
For the first part, we make use of the following Lemma from \cite{candes2007sparsity,myp}.
\begin{lemma}
Suppose $m_2$ rows are sampled uniformly at random (without replacement) from the matrix $\bL$ with rank $r$. If
\begin{eqnarray}
m_2 \ge r\eta_u^2  \max \left( c_1 \log r , c_2 \log\left(\frac{3}{\delta}\right) \right),
\label{eq27}
\end{eqnarray}
then the selected rows of the matrix $\bL$ span the row space of $\bL$ with probability at least $(1-\delta)$, where $c_1$ and $c_2$ are numerical constants.
\label{lm on paper}
\end{lemma}
The matrices $\bL_s$ and $\bL$ have the same column space. Thus, if $m_2$ satisfies (\ref{eq27}), the rank of $\mathbf{\Phi}\bL_s$ is equal to the rank of $\bL_s$ with probability at least $1 - \delta$.
Now we prove the second part. Assume the first part is satisfied, i.e., the rank of $\bL_s^{\phi}$ is equal to $r$. It is easy to show that since ${\bU_s^{\phi}}^\perp$ is an orthonormal matrix, then the  elements of matrix (\ref{eq proj}) are zero-mean independent normal random variables with equal variance. In order to show that the rank of (\ref{eq proj}) is equal to $k$, we make use of the following lemma from \cite{lamport14,davidson2001local}.

\begin{lemma}
Let $\bA$ be an $N \times n $ matrix whose entries are independent standard normal variables. Then for every $t \ge \sqrt{2 \log 2/\delta} $,
\begin{eqnarray}
\sqrt{N} - \sqrt{n} - t \leq \sigma_{min} (\bA) \leq \sigma_{max} (\bA) \leq \sqrt{N} + \sqrt{n} + t \:
\end{eqnarray}
with probability at least $1 - \delta$, where $\sigma_{min} (\bA)$ and $\sigma_{max} (\bA)$ are the minimum and maximum singular values of $\bA$. %respectively.
\label{lemma: singulars}
\end{lemma}

Define $\bZ$ as the non-zero columns of the matrix in (\ref{eq proj}). Based on Lemma (\ref{lemma: singulars}), to prove that the rank of $\bZ$ is equal to $k$ with probability at least $1 - \delta$, it suffices to have
$$\sqrt{m_2 - r} - \sqrt{q}  > \sqrt{2 \log 2/\delta}.$$
\smallbreak
\noindent\textbf{Proof of Lemma \ref{lemma row 2}}\\
Similar to the proof of Lemma \ref{lemma row 1}, we can guarantee that if $m_2$ satisfies inequality (\ref{m_2 row 2}), then the rank of  $\mathbf{\Phi}\bL$ is equal to the rank of $\bL$ with probability $1- \delta$.

Suppose $\bc$ is a non-zero column of $\bC_s$. Similar to the analysis provided in the proof of Lemma \ref{lemma row 1}, if
$$\sqrt{m_2 - r} - 1  > \sqrt{2 \log 2/\delta} \: , $$
then $\mathbf{\Phi} \bc$ do not lie in the column space of $\mathbf{\Phi}\bL$ with probability at least $1- \delta$. Thus, if
$$m_2 > r + 1 + 2 \log 2q /\delta + 2\sqrt{2 \log 2q /\delta} \: , $$
then the non-zero columns of $\mathbf{\Phi} \bC_s$ do not lie in the column space of $\mathbf{\Phi} \bL$ with probability at least $1 - \delta$.
\smallbreak

\noindent\textbf{ Proof of Theorem \ref{theorem 1}}\\
Algorithm 1 with RED recovers the exact subspace if: \\
\textbf{[I]} The inliers of $\bD_s$ span the column space of $\bL$, and each inlier of $\bD_s$ lies in the span of the other inlier columns of $\bD_s$. \\
\textbf{[II]} If $\bD_s$ contains $k$ outlying columns, the rank of $\mathbf{\Phi} \bD_s$ is equal to $r+k$.

Lemma \ref{lm4} and Lemma \ref{lm6} establish a sufficient condition for $m_1$ to guarantee \textbf{[I]} whp. Given Assumption 1, the rank of $\bD_s$ is equal to $r+k$.  Lemma \ref{lemma: embd 1} provides a sufficient condition for $m_2$ to ensure that the rank of $\mathbf{\Phi}\bD_s$ is equal to the rank of $\bD_s$ whp, i.e., \textbf{[II]} is guaranteed whp.
 In addition, Lemma \ref{lm7} provides an upper-bound on the number of sampled outliers. Therefore, according to Lemma \ref{lm4}, Lemma \ref{lm6}, Lemma \ref{lemma: embd 1} and Lemma \ref{lm7}, if (\ref{suff thorem 1}) is satisfied, Algorithm 1 with RED recovers the correct subspace with probability at least $1 - 4 \delta$.

In addition, similar to the analysis provided in the proof of Lemma \ref{lemma: embd 2}, if
\begin{eqnarray}
\begin{aligned}
m_2 \ge \frac{(r+1) \log (42 \sqrt{2}) + \log K + \log \frac{2}{\delta}}{f (\frac{1}{{2}})}   \: ,
\end{aligned}
\end{eqnarray}
then the non-zero columns of $\mathbf{\Phi}\bC$ do not lie in the column space of $\mathbf{\Phi}\bL$ with probability at least $1 - \delta$. Thus, if the subspace is learned correctly, (\ref{outlier_detect 1}) identifies the outlying columns correctly with probability at least $1 - \delta$.
\smallbreak

\noindent\textbf{Proof of Theorem \ref{Theorem 2}}\\
The proof of Theorem \ref{Theorem 2} is similar to the proof of Theorem \ref{theorem 1} but we need to make use of Lemma \ref{lemma row 1} (instead of Lemma \ref{lemma: embd 1}) to guarantee \textbf{[II]} whp. Therefore, according to Lemma \ref{lm4}, Lemma \ref{lm6}, Lemma \ref{lemma row 1} and Lemma \ref{lm7}, if the requirements of Theorem \ref{Theorem 2} are satisfied, Algorithm 1 with RRD recovers the correct subspace with probability at least $1 - 5 \delta$.
In addition, similar to the analysis provided in the proof of Lemma \ref{lemma row 2}, if
\begin{eqnarray}
\begin{aligned}
   m_2 \ge
 r+ 1 +2 \log \frac{2 K}{\delta} + \sqrt{8 \: \log \frac{2 K}{\delta} }  \Bigg] \:,
\end{aligned}
\end{eqnarray}
then the non-zero columns of $\mathbf{\Phi}\bC$ do not lie in the column space of $\mathbf{\Phi}\bL$ with probability at least $1 - \delta$. Thus, if the subspace is learned correctly, (\ref{outlier_detect 1}) identifies the outlying columns correctly with probability at least $1 - \delta$.
\smallbreak

\noindent\textbf{Proof of Theorem \ref{Theorem 3}}\\
In order to guarantee that Algorithm 2  recovers the exact subspace, we have to ensure that\\
\textbf{(a)} The columns of $\bL_s$ span the column space of $\bL$. \\
\textbf{(b)} Requirement 2 is satisfied.\\
\textbf{(c)} The optimization problem (\ref{convex_in_alg}) yields correct decomposition, i.e., the column space of $\hat{\bL}_s^{\phi}$ is equal to the column space of $\mathbf{\Phi}\bL$ and the non-zero columns of  $\hat{\bC}_s^{\phi}$ and $\mathbf{\Phi} \bC_s$ are at the same locations.

\noindent\textbf{Guarantee for (a)}:\\
 It suffices to show that the rank of $\bV^T \bS$ is equal to $r$.
According to the proof of Lemma \ref{lm4}, if we set
\begin{eqnarray}
m_1 \ge 10 \: r \mu_v \log \frac{2 r}{\delta}
\label{eq31n}
\end{eqnarray}
then,
\begin{eqnarray}
\mathbb{P} \left[ \frac{N_2}{m_1} \Bigg \|  \bV^T \bS \bS^T \bV  - \frac{m_1}{N_2} \bI  \Bigg  \| > \frac{1}{2} \right] \leq \delta \:.
\label{eq32n}
\end{eqnarray}
If $\sigma_1$ and $\sigma_r$ denote the largest and smallest singular values of $\bS^T \bV$, respectively, then
\begin{eqnarray}
\frac{m_1}{2 N_2} \leq \sigma_1^2 \leq \sigma_r^2 \leq \frac{3 m_1}{2 N_2}
\label{eq33n}
\end{eqnarray}
with probability at least $1- \delta$. Accordingly, the matrix $\bV^T \bS$ is a full rank matrix with probability at least $1- \delta$. In addition, we study the row space coherency of matrix $\bL_s$ since it is used to derive the  guarantee for \textbf{(c)}. The projection of the standard basis onto the row space of $\bL_{s}$ can be written as
\begin{eqnarray}
\begin{aligned}
&\underset{i}{\max} \| \textbf{P}_{\bS^T \bV} \be_i \|_2^2 = \underset{i}{\max} \| \bS^T \bV (\bV^T \bS \bS^T \bV)^{-1} \bV^T \bS \be_i \|_2^2 \\
& \leq \underset{j}{\max} \| \bS^T \bV (\bV^T \bS \bS^T \bV)^{-1} \bV^T  \be_j \|_2^2\\
&\leq \| \bS^T \bV (\bV^T \bS \bS^T \bV)^{-1} \|^2 \| \bV^T  \be_j \|_2^2 \\
& \leq \frac{\mu_v r}{N_2} (\frac{\sigma_1^2}{\sigma_r^4})= \frac{\mu_v r}{N_2} \frac{6 N_2}{m_1}=\frac{6 \mu_v r}{m_1}
\end{aligned}
\label{eq46}
\end{eqnarray}
where $ (\bS^T \bV (\bV^T \bS \bS^T \bV)^{-1} \bV^T \bS)$ is the projection matrix onto the column space of $\bS^T \bV$. The first inequality follows from the fact that $\{ \bS \be_i \}_{i=1}^{m_1}$ is a subset of $\{ \be_j \}_{j=1}^{N_2}$. The second inequality follows from Cauchy-Schwarz inequality and the third inequality follows from (\ref{eq33n}).

\smallbreak
\noindent\textbf{Guarantee for (b)}:\\
Suppose that \textbf{(a)} is true. If $\bq$ is the number of outliers of $\bD_s$, Lemma \ref{lemma: embd 2} provides a sufficient condition for $m_2$ (inequality (\ref{eq:  suffi m_2 2})) to guarantee that these requirements are satisfied.
\smallbreak

\noindent\textbf{Guarantee for (c)}:\\
Suppose \textbf{(a)} and \textbf{(b)} are satisfied.
First, let us review the theoretical result provided in \cite{lamport10} which supports the performance of the convex algorithm (\ref{eq6}).
\begin{lemma}
Suppose $\bD$ follows Data model 1 and define $\bL^{*}$ and $\bC^{*}$ as the optimal point of (\ref{eq6}). If
\begin{eqnarray}
K \leq \left(  \frac{1}{1 + (121/9) r \gamma }    \right) N_2 \quad \text{and} \quad \lambda = \frac{3}{7 \sqrt{K}},
\label{eq47}
\end{eqnarray}
then the column space of $\bL^{*}$ is equal to the column space of $\bL$ and the location of non-zero columns of $\bC^{*}$ indicate the location of non-zero columns of $\bC$.
\label{lm9}
\end{lemma}
The matrix $\bD^{\phi}_s$ can be expressed as $\bD^{\phi}_s = \bL^{\phi}_s + \bC^{\phi}_s$, where $\bL^{\phi}_s = \mathbf{\Phi}\bL \bS$ and $\bC^{\phi}_s = \mathbf{\Phi}\bC \bS$. If the rank of $\mathbf{\Phi} \bL$ is equal to $r$, then $\bL_s$ and $\bL_s^{\phi}$ have the same row space. Thus, if $\bV_s^{\phi}$ is an orthonormal basis for the row space of $\bL^{\phi}_s$, then from (\ref{eq46})
\begin{eqnarray}
\underset{i}{\max} \|(\bV_s^{\phi})^T \be_i\|_2^2 \leq \frac{6 r \mu_v }{m_1}.
\label{eq54}
\end{eqnarray}
Define $n_{L^{\phi}_s}$ as the   number of non-zero columns of $\bL^{\phi}_s$. Therefore,
\begin{eqnarray}
\underset{i}{\max} \|(\bV^{\phi}_s)^T \be_i\|_2^2 \leq \frac{6 r \mu_v }{m_1} \leq  \frac{6 r \mu_v }{n_{L^{\phi}_s}} .
\label{eq55}
\end{eqnarray}
Suppose $m_1 = \zeta \frac{N_2}{N_2^{'}}$.
According to Lemma \ref{lm7}, if
$
\zeta \ge 3 g^2 \frac{K}{N_2^{'}} \log \frac{2}{\delta},
$
then the number of outlying columns of $\bD^{\psi}_s$ is less than or equal to
\begin{eqnarray}
\zeta \left( \frac{K}{N_2^{'}} + \frac{1}{g} \right)
\label{eq57}
\end{eqnarray}
with probability at least $(1 - \delta)$ where $g$ can be any number greater than one. Therefore, if
$
m_1 \ge \frac{N_2}{N_2^{'}} \left( 3 g^2 \frac{K}{N_2} \log \frac{2}{\delta} \right)
$
 and
\begin{eqnarray}
g \ge \frac{N_2^{'}}{N_2} \left( 1 + 6 r \mu_v (121/9) \right) \: ,
\label{eq59}
\end{eqnarray}
then according to Lemma \ref{lm9}, the column space of $\hat{\bL}_s^{\phi}$ is equal to the column space of $\mathbf{\Phi}\bL$ and the non-zero columns of  $\hat{\bC}_s^{\phi}$ and $\mathbf{\Phi} \bC_s$ are at the same locations provided that
\begin{eqnarray}
\frac{K}{N_2^{'}} \leq \frac{g \frac{N_2}{N_2^{'}} - (1 + 6 r \mu_v (121/9))}{g (1+6 r \mu_v (121/9))} .
\label{eq60}
\end{eqnarray}

%Define
%$$
%\zeta = \max \left( 3 g^2 \frac{K}{N_2^{'}} \log \frac{2}{\delta} \: , \frac{N_2^{'}}{N_2} 10\: r \mu_v  \log \frac{2r}{\delta} \right) \: .
%$$
%Therefore, $m_1 \ge \frac{N_2}{N_2^{'}}\zeta$  and (\ref{eq:  suffi m_2 2}), (\ref{eq59}) and ((\ref{eq60})) are satisfied, where $\bq$ is equal to (\ref{eq57}), then Algorithm 2 extract the correct subspace with probability at least $1 - 3 \delta$.
Therefore, if the requirements of Theorem \ref{Theorem 3} are satisfied, Algorithm 2 with RED extracts the exact subspace with probability at least $1 - 3 \delta$. In addition, according to the analysis provided in the proof of Lemma \ref{lemma: embd 2}, if $m_2$ satisfies the requirement of Theorem \ref{Theorem 3}, then the columns of $\mathbf{\Phi} \bC_s$ do not lie in the column space of $\mathbf{\Phi} \bL$, and the non-zero columns of $\mathbf{\Phi} \bC$ do not lie in the column space of $\mathbf{\Phi} \bL$ whp, i.e., if the exact subspace is retrieved, (\ref{outlier_detect 1}) identifies the outlying columns correctly whp.
\smallbreak

\noindent\textbf{Proof of Theorem \ref{theorem 4}}\\
The proof of Theorem \ref{theorem 4} is similar to the proof of Theorem \ref{Theorem 3}. But, we use Lemma \ref{lemma row 2} to establish a sufficient condition on $m_2$ to guarantee \textbf{(b)}. In addition, according to the analysis in the proof of Lemma \ref{lemma row 2}, if $m_2$ satisfies the requirement of Theorem \ref{theorem 4}, not only is Requirement 2 satisfied whp,
but also the non-zero columns of $\mathbf{\Phi} \bC$ do not lie in the column space of $\mathbf{\Phi} \bL$ whp, i.e., (\ref{outlier_detect 1}) identifies the outlying columns correctly whp in case of exact subspace recovery.
%\nocite{*}
%\bibliographystyle{IEEEbib}
%{\small
%\bibliography{bibfile}%}

\bibliographystyle{IEEEtran}
\bibliography{IEEEabrv,bibfile}

%\bibliography{bibfile, refs}

%\bibliography{strings,refs}

\end{document}

%% file: preamble.tex
\usepackage[mathscr]{eucal}
\usepackage[cmex10]{amsmath}
\usepackage{epsfig,epsf,psfrag}
\usepackage{amssymb,amsmath,amsthm,amsfonts,latexsym}
\usepackage{amsmath,graphicx,bm,xcolor,url}
\usepackage[caption=false]{subfig} 
\usepackage{fixltx2e}%ordering of single and double column floats
\usepackage{array}%array and tabular environments
\usepackage{verbatim}
\usepackage{bm}
\usepackage{algorithmic, cite}
\usepackage{algorithm}
\usepackage{verbatim}
\usepackage{textcomp}
\usepackage{mathrsfs}
\usepackage{epstopdf}
\catcode`~=11 \def\UrlSpecials{\do\~{\kern -.15em\lower .7ex\hbox{~}\kern .04em}} \catcode`~=13 

\allowdisplaybreaks[3]

\newcommand{\calN}{\mathcal{N}}
\newcommand{\calO}{\mathcal{O}}

\newcommand{\calS}{\mathcal{S}}

\newcommand{\ba}{\mathbf{a}}
\newcommand{\bA}{\mathbf{A}}

\newcommand{\bB}{\mathbf{B}}
\newcommand{\bc}{\mathbf{c}}
\newcommand{\bC}{\mathbf{C}}
\newcommand{\bd}{\mathbf{d}}
\newcommand{\bD}{\mathbf{D}}
\newcommand{\be}{\mathbf{e}}

\newcommand{\bG}{\mathbf{G}}

\newcommand{\bH}{\mathbf{H}}

\newcommand{\bI}{\mathbf{I}}

\newcommand{\bL}{\mathbf{L}}

\newcommand{\bN}{\mathbf{N}}

\newcommand{\bq}{\mathbf{q}}
\newcommand{\bQ}{\mathbf{Q}}

\newcommand{\bs}{\mathbf{s}}
\newcommand{\bS}{\mathbf{S}}

\newcommand{\bT}{\mathbf{T}}

\newcommand{\bU}{\mathbf{U}}
\newcommand{\bv}{\mathbf{v}}
\newcommand{\bV}{\mathbf{V}}

\newcommand{\bX}{\mathbf{X}}

\newcommand{\bz}{\mathbf{z}}
\newcommand{\bZ}{\mathbf{Z}}

\DeclareMathAlphabet{\mathbsf}{OT1}{cmss}{bx}{n}
\DeclareMathAlphabet{\mathssf}{OT1}{cmss}{m}{sl}% slanted sans serif

\DeclareSymbolFont{bsfletters}{OT1}{cmss}{bx}{n}  
\DeclareSymbolFont{ssfletters}{OT1}{cmss}{m}{n}
\DeclareMathSymbol{\bsfGamma}{0}{bsfletters}{'000}
\DeclareMathSymbol{\ssfGamma}{0}{ssfletters}{'000}
\DeclareMathSymbol{\bsfDelta}{0}{bsfletters}{'001}
\DeclareMathSymbol{\ssfDelta}{0}{ssfletters}{'001}
\DeclareMathSymbol{\bsfTheta}{0}{bsfletters}{'002}
\DeclareMathSymbol{\ssfTheta}{0}{ssfletters}{'002}
\DeclareMathSymbol{\bsfLambda}{0}{bsfletters}{'003}
\DeclareMathSymbol{\ssfLambda}{0}{ssfletters}{'003}
\DeclareMathSymbol{\bsfXi}{0}{bsfletters}{'004}
\DeclareMathSymbol{\ssfXi}{0}{ssfletters}{'004}
\DeclareMathSymbol{\bsfPi}{0}{bsfletters}{'005}
\DeclareMathSymbol{\ssfPi}{0}{ssfletters}{'005}
\DeclareMathSymbol{\bsfSigma}{0}{bsfletters}{'006}
\DeclareMathSymbol{\ssfSigma}{0}{ssfletters}{'006}
\DeclareMathSymbol{\bsfUpsilon}{0}{bsfletters}{'007}
\DeclareMathSymbol{\ssfUpsilon}{0}{ssfletters}{'007}
\DeclareMathSymbol{\bsfPhi}{0}{bsfletters}{'010}
\DeclareMathSymbol{\ssfPhi}{0}{ssfletters}{'010}
\DeclareMathSymbol{\bsfPsi}{0}{bsfletters}{'011}
\DeclareMathSymbol{\ssfPsi}{0}{ssfletters}{'011}
\DeclareMathSymbol{\bsfOmega}{0}{bsfletters}{'012}
\DeclareMathSymbol{\ssfOmega}{0}{ssfletters}{'012}

\DeclareMathOperator{\spn}{span}

\DeclareMathOperator{\rank}{rank}

\newtheorem{theorem}{Theorem} 
\newtheorem{lemma}[theorem]{Lemma}

\newtheorem{definition}{Definition}

\newtheorem{remark}{Remark}

\newtheorem{assumption}{Assumption}
\newtheorem{data model}{Data Model}
\newtheorem{requirement}{Requirement}

\newcommand{\qednew}{\nobreak \ifvmode \relax \else
      \ifdim\lastskip<1.5em \hskip-\lastskip
      \hskip1.5em plus0em minus0.5em \fi \nobreak
      \vrule height0.75em width0.5em depth0.25em\fi}